\documentclass{article}
\usepackage[preprint]{neurips_2025}
\usepackage{algorithm}
\usepackage{algpseudocode}
\usepackage{multirow}

\usepackage{float}

\usepackage{wrapfig}

\usepackage{svg}

\usepackage[utf8]{inputenc} 
\usepackage{xcolor}
\usepackage{amsmath}

\usepackage[T1]{fontenc}    
\usepackage{hyperref}
\hypersetup{
    colorlinks=true,
    linkcolor={rgb,255:red,0;green,102;blue,204},
    citecolor={rgb,255:red,0;green,102;blue,204},
    urlcolor={rgb,255:red,255;green,0;blue,204}  
}
\usepackage{url}            
\usepackage{booktabs}       
\usepackage{amsfonts}       
\usepackage{nicefrac}       
\usepackage{microtype}      
\usepackage{etoolbox}
\usepackage{caption} 
\usepackage{adjustbox}

\usepackage{float}
\usepackage{graphicx}

\usepackage{titlesec}
\usepackage{titletoc}
\usepackage{tocloft}

\title{Semantic and Visual Crop-Guided Diffusion Models for Heterogeneous Tissue Synthesis in Histopathology}

\author{
Saghir Alfasly\quad Wataru Uegami\quad MD Enamul Hoq\quad Ghazal Alabtah\quad H.R. Tizhoosh\thanks{Corresponding author}  \\
KIMIA Lab, Department of Artificial Intelligence and Informatics, Mayo Clinic, Rochester, MN 55901\\
\texttt{\{Alfasly.Saghir, Tizhoosh.Hamid\}@mayo.edu} \\
\texttt{\href{https://kimialabmayo.github.io/hetero_tissue_diffuse_page/}{\textcolor[RGB]{0,102,204}{https://kimialabmayo.github.io/hetero\_tissue\_diffuse\_page/}}
}}

\begin{document}
\maketitle

\begin{abstract}
Synthetic data generation in histopathology faces unique challenges: preserving tissue heterogeneity, capturing subtle morphological features, and scaling to unannotated datasets. We present a latent diffusion model that generates realistic heterogeneous histopathology images through a novel dual-conditioning approach combining semantic segmentation maps with tissue-specific visual crops. Unlike existing methods that rely on text prompts or abstract visual embeddings, our approach preserves critical morphological details by directly incorporating raw tissue crops from corresponding semantic regions. For annotated datasets (i.e., Camelyon16, Panda), we extract patches ensuring $20-80\%$ tissue heterogeneity. For unannotated data (i.e., TCGA), we introduce a self-supervised extension that clusters whole-slide images into 100 tissue types using foundation model embeddings, automatically generating pseudo-semantic maps for training. Our method synthesizes high-fidelity images with precise region-wise annotations, achieving superior performance on downstream segmentation tasks. When evaluated on annotated datasets, models trained on our synthetic data show competitive performance to those trained on real data, demonstrating the utility of controlled heterogeneous tissue generation. In quantitative evaluation, prompt‐guided synthesis reduces Fréchet Distance by up to $6\times$ on Camelyon16 (from $430.1$ to $72.0$) and yields $2-3×$ lower FD across Panda and TCGA. Downstream DeepLabv3+ models trained solely on synthetic data attain test IoU of $0.71$ and $0.95$ on Camelyon16 and Panda, within $1-2\%$ of real‐data baselines ($0.72$ and $0.96$). By scaling to $11,765$ TCGA whole‐slide images without manual annotations, our framework offers a practical solution for an urgent need for generating diverse, annotated histopathology data, addressing a critical bottleneck in computational pathology.
\end{abstract}

\section{Introduction}

Histopathology image analysis forms the cornerstone of cancer diagnosis, yet remains constrained by data scarcity, laborious annotation processes, and privacy concerns \cite{campanella2019clinical, komura2018machine,louis2014digital, abels2019computational, kaissis2020secure}. While generative AI has transformed natural image synthesis, its application to histopathological imaging in digital pathology faces unique challenges due to the complex, multi-scale architecture of biological tissues and the critical importance of preserving diagnostically relevant features \cite{wei2019generative, gadermayr2019generative, afshari2023single}. Most of the current approaches have predominantly focused on generating homogeneous tissue types using text-based prompting systems, which introduce significant interobserver variability and limit clinical utility, particularly problematic in a domain where expert agreement is already inconsistent \cite{elmore2015diagnostic, tizhoosh2021searching}.

The evolution from GAN-based methods to diffusion models has improved image quality and training stability \cite{dhariwal2021diffusion, ho2020denoising}, but existing frameworks fail to account for the heterogeneous nature of real-world histopathology samples. Levine et al. \cite{levine2020synthesis} demonstrated that GANs could generate images indistinguishable from real histopathology samples, while Moghadam et al. \cite{moghadam2023morphology} marked the transition to diffusion models with superior image quality and training stability. Despite these advances, clinical specimens typically contain multiple tissue types and pathological features within a single slide, requiring region-specific control over generation parameters. This limitation severely restricts the utility of synthetic data for developing robust diagnostic algorithms that must recognize complex patterns in diverse tissue regions.

\begin{figure}
    \centering
    \includegraphics[width=0.95\linewidth]{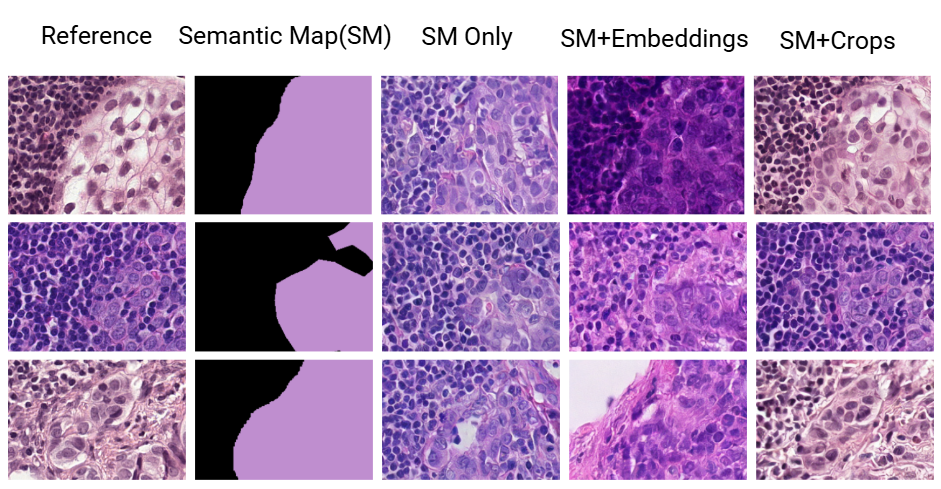}
\caption{\textbf{Comparison of reference and synthetic Camelyon16 patches using three conditioning schemes.} From left to right: original histopathology patch;  binary tumor–normal mask;  generated image with a semantic segmentation map conditioning only;  synthetic image by combining conditions of semantic maps and abstract embeddings; synthetic output of our model conditioned on the semantic map and tissue-specific visual crop prompts. The crop-guided generation recovers fine morphological details and staining heterogeneity more faithfully than embedding-based conditioning.}
    \label{fig:illustration}
\end{figure}
Further complicating this landscape is the tension between the global architecture of the tissue and the local cellular details. Traditional generative approaches struggle to maintain consistency across multiple magnification levels while preserving the fine-grained morphological features critical for diagnosis \cite{quiros2021pathology, tellez2019quantifying, afshari2023single}. Recent work, URCDM \cite{cechnicka2024urcdm}, has addressed this through cascaded diffusion models generating images at multiple resolutions simultaneously, while \emph{DiffInfinite} \cite{aversa2024diffinfinite} enables arbitrary-size image synthesis with preserved long-range structural correlations. However, this multiscale challenge remains particularly acute when attempting to synthesize realistic transitions between different tissue types, a capability essential for training segmentation algorithms and supporting differential diagnosis.

\begin{figure}
    \centering
    \includegraphics[width=\linewidth]{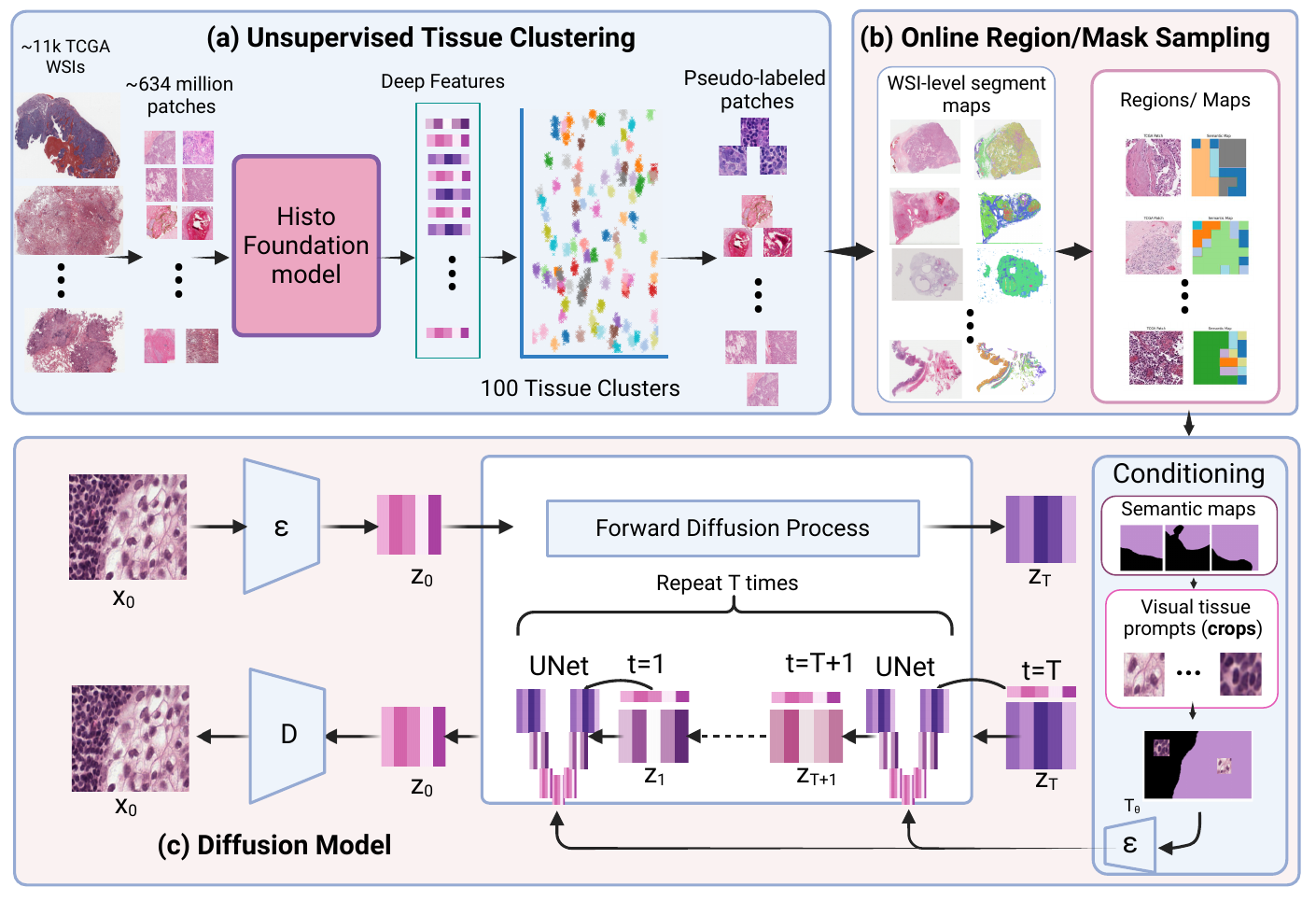}
    \caption{\textbf{Schematic overview of the HeteroTissue-Diffuse framework for heterogeneous tissue synthesis in histopathology.} (a) Unsupervised Tissue Clustering: For unannotated datasets (TCGA), approximately $11,765$ whole-slide images (WSIs) are processed to extract 634,435,134 million patches. A histopathology foundation model extracts deep features, which are used to cluster patches into 100 distinct tissue types, creating pseudo-labeled data. (b) Online Region/Mask Sampling: The pseudo-labeled patches are used to generate WSI-level segmentation maps and regional masks for conditioning the diffusion model. (c) Diffusion Model: Our dual-conditioning approach combines semantic maps with visual tissue prompts (crops) to guide the latent diffusion process. The model encodes the input image to the latent space, applies forward diffusion to create a noisy latent, then reverses this process with UNet denoising conditioned on both semantic maps and visual tissue exemplars. For annotated datasets (Camelyon16 \cite{bejnordi2017diagnostic} and Panda~\cite{bulten2022artificial}), only component (c) is used with their existing semantic maps.
    }\vspace{-0.5em}
    \label{fig:mainfig}
\end{figure}

Our work addresses these challenges by introducing a visually prompted latent diffusion model designed specifically for heterogeneous tissue synthesis in histopathology. Unlike text-guided approaches, our framework leverages spatial masks and visual exemplars to provide fine-grained control over region-specific generation. Previous work, in the literature, NASDM \cite{shrivastava2023nasdm} and subsequent extensions by Konz et al. \cite{konz2024segguided} and Xu et al. \cite{xu2024topocellgen} demonstrated the value of region-guided generation and semantic instance masks.  Our method expands this concept to enable the synthesis of complex, multi-tissue samples with realistic transitions and clinically relevant features as shown in Figure \ref{fig:illustration}. Our approach significantly reduces the annotation burden while maintaining high fidelity to real-world pathological presentations \ref{fig:mainfig}. The main contributions of our work include:
\vspace{-0.7em}
\begin{itemize}
    \item \textbf{New Dual-Conditioning Architecture for Histopathology Synthesis.}
We developed a unique visual crop-guided diffusion model that combines semantic maps with raw tissue exemplars, preserving critical morphological features (nuclear texture, staining patterns, cellular structure) that are lost in text-based or embedding-based approaches. This enables precise region-specific control while maintaining authentic tissue appearance for heterogeneous sample generation. \vspace{-0.3em}
\item \textbf{Self-Supervised Framework for Unannotated Whole-Slide Images.}
We introduced a scalable solution for the TCGA dataset (11,765 WSIs) that automatically discovers and clusters 100 distinct tissue phenotypes without manual annotation. This democratizes access to diverse synthetic data across 33 cancer types while preserving patient privacy and addressing critical data scarcity in computational pathology. \vspace{-0.3em}
\item \textbf{Comprehensive Multi-Modal Validation Framework.}
We established a rigorous evaluation pipeline combining quantitative metrics (Fréchet distance across 8 foundation model encoders) with downstream task performance (segmentation IoU scores). Most notably, our blinded assessment by certified pathologists using a 5-point Likert scale across multiple quality criteria revealed that our synthetic images were indistinguishable from real samples, with one pathologist commenting: "The generated images tended to have equal or higher quality than the real images."
\end{itemize}
\vspace{-0.5em}

Through rigorous evaluation involving both quantitative metrics (including the Fréchet Inception Distance \cite{borji2022pros}) and expert pathologist evaluation, we demonstrate that our approach generates histopathology images that are indistinguishable from real samples while providing \emph{unprecedented control over tissue composition}. Synthetic datasets generated using our method effectively augment or replace real data in training diagnostic models, addressing the critical issue of data scarcity while preserving patient privacy.\vspace{-0.2em}

By enabling the generation of diverse, annotated histopathology datasets without requiring patient data sharing, our framework represents a significant step toward more equitable and robust AI development in computational pathology \cite{larson2021ethics}. This capability is particularly valuable for rare cancer types and underrepresented populations, potentially democratizing access to high-quality training data between institutions, regardless of their size or resources.\vspace{-0.5em}
\section{Related Work}\vspace{-0.5em}
\textbf{Generative Models in Histopathology.}
Visual generative models have evolved from early GANs~\cite{goodfellow2014generative} to sophisticated diffusion models~\cite{ho2020denoising}, with recent advances like ControlNet~\cite{zhang2023adding} enabling fine-grained control. Adapting these to histopathology presents unique challenges due to the complexity of the tissue and the diagnostic significance of subtle morphological features. Although early GANs demonstrated feasibility~\cite{levine2020synthesis,quiros2021pathology,afshari2023single}, recent diffusion-based approaches show superior quality~\cite{moghadam2023morphology}. Domain-specific methods ~\cite{ktena2024generative, kather2022pan, zhao2020data,cechnicka2024urcdm} have emerged, with URCDM~\cite{cechnicka2024urcdm} addressing multi-resolution synthesis and enabling arbitrary-size generation ~\cite{aversa2024diffinfinite, tschuchnig2022generative}. However, most approaches generate homogeneous tissue types, limiting their utility for training diagnostic models that require heterogeneous tissue representations, a limitation that our framework specifically addresses.
\vspace{-0.2em}

\textbf{Conditioning Mechanisms for Histopathology Synthesis.}
Existing conditioning approaches fall into three categories, each with significant limitations. Unconditioned models~\cite{wei2019generative,gadermayr2019generative} produce realistic images but lack control over tissue types and pathological features, severely limiting their utility for training task-specific models. Metadata-guided, text-guided, or mask-guided models~\cite{drexlin2025medi, yellapragada2024pathldm, xu2024topocellgen, carrillo2025generation} suffer from interobserver variability, as documented by Elmore et al.~\cite{elmore2015diagnostic} who found substantial disagreement among pathologists (kappa values as low as 0.48). Visual embedding or RNA-seq embedding approaches~\cite{osorio2024latent, ciga2022self,zhang2022contrastive} avoid text ambiguity, but introduce lossy transformations that can obscure critical diagnostic features. Even domain-specific embeddings suffer from information loss during dimensionality reduction. Our approach circumvents these limitations by directly conditioning on \emph{real tissue crops combined with semantic maps}, preserving original visual characteristics without intermediate representations.
\vspace{-0.2em}

\textbf{Semantic Map-Based Generation.}
Recent works have explored semantic map conditioning for precise spatial control. Shrivastava and Fletcher~\cite{shrivastava2023nasdm} pioneered this with NASDM, while Konz et al.~\cite{konz2024segguided} extended it through random mask ablation. Although spatially accurate, these approaches typically focus on single tissue types or cellular structures rather than heterogeneous tissue architectures. Our dual-condition mechanism combines semantic maps with visual crops, enabling the synthesis of diverse tissue compositions while maintaining both spatial accuracy and morphological fidelity, essential for generating comprehensive segmentation datasets.
\vspace{-0.2em}

\textbf{Large-Scale Synthesis and Evaluation.}
Whole-slide image (WSI) synthesis presents unique challenges due to gigapixel resolution and structural dependencies. Cechnicka et al.~\cite{cechnicka2024urcdm} and Aversa et al.~\cite{aversa2024diffinfinite} addressed scale issues through cascaded models and infinite tiling, respectively, but struggled to maintain segmentation accuracy with visual characteristics. Our self-supervised framework in this paper, trained on TCGA's 11,765 diagnostic WSIs, generates 100 distinct tissue types while preserving region-specific control through dual conditioning.

Evaluation of synthetic histopathology requires specialized metrics beyond standard FID and IS, which use networks pretrained on natural images. Domain-specific alternatives include Fréchet Distance and Topological Fréchet Distance~\cite{xu2024topocellgen}, complemented by the evaluation of expert pathologists~\cite{srinidhi2021deep}. Our comprehensive validation combines these metrics with downstream task performance, the ultimate measure of the utility of synthetic data.
\vspace{-0.2em}

\textbf{Our work advances the field through three key innovations.} First, we avoid intermediate representations that compromise fidelity by directly conditioning on visual crops from real histopathology images combined with spatial semantic maps. Unlike semantic-only methods~\cite{shrivastava2023nasdm,konz2024segguided}, text-based approaches~\cite{rombach2022high}, or embedding-based techniques~\cite{huang2023t2i,radford2021learning}, our approach preserves tissue-specific attributes (i.e., texture, cellular morphology, and staining patterns) that abstract representations lose. This direct visual conditioning enables the synthesis of realistic heterogeneous samples with precise spatial control, which is essential to train robust diagnostic models. Second, our self-supervised extension to TCGA democratizes access to diverse synthetic data across cancer types, generating 100 distinct tissue phenotypes without manual annotation while addressing critical data scarcity and preserving patient privacy~\cite{larson2021ethics}. Third, we establish comprehensive validation through rigorous metrics combining expert pathologist assessment with quantitative measures (Fréchet distance, precision, recall, F1 score), and demonstrate utility through downstream segmentation tasks that confirm the high quality of our generated annotated data.\vspace{-0.5em}
\vspace{-0.5em}
\section{Method}

Our approach addresses the fundamental challenge in histopathology synthesis: generating realistic heterogeneous tissue images with precise region-based control while preserving morphological fidelity. We achieve this through a novel dual-conditioning latent diffusion model that combines semantic maps with tissue-specific visual crops. 

\subsection{Preliminaries: Latent Diffusion Models}

\begin{wrapfigure}{r}{0.50\textwidth}
\vspace{-20pt}
\tiny
\begin{minipage}{\linewidth}
\captionof{algorithm}{Heterogeneous Patch Sampling for Annotated Data}
\label{alg:annotated_sampling}
\vspace{2pt}
\fbox{
\begin{minipage}{\dimexpr\linewidth-2\fboxsep-2\fboxrule}
\begin{algorithmic}[1]
\Require Dataset $\mathcal{D}$ with patches and segmentation masks
\Require Tissue ratio bounds $[r_{\min}, r_{\max}] = [0.2, 0.8]$
\Require Crop size range $[d_{\min}, d_{\max}] = [50, 200]$
\Ensure Training sample $(x, c)$
\Function{SampleHeterogeneousPatch}{$\mathcal{D}$}
    \State $(x, M) \gets$ SelectPatch($\mathcal{D}$) where tissue ratios $\in [r_{\min}, r_{\max}]$
    \State $K \gets$ number of tissue classes
    \State Initialize $C \gets$ zeros($H, W, 3K$)
    
    \For{$k = 1$ to $K$}
        \State $C[:,:,(k-1)] \gets M[:,:,k]$ \Comment{Semantic channel}
        
        \If{$\sum M[:,:,k] > 0$} \Comment{Class $k$ present}
            \State $d \gets$ RandomInt($d_{\min}, d_{\max}$)
            \State $p_k, (r, c) \gets$ ExtractSquareCrop($x, M[:,:,k], d$)
            \State $p_k \gets$ Augment($p_k$) \Comment{Optional rotation/flip}
            \State $C_k \gets$ zeros($H, W, 3$)
            \State $C_k[r:r+d, c:c+d, :] \gets p_k$
            \State $C[:,:,K+(k-1)*3:K+k*3] \gets C_k$
        \EndIf
    \EndFor
    
    \State \Return $(x, c)$
\EndFunction
\end{algorithmic}
\end{minipage}
}
\end{minipage}
\vspace{-20pt}
\end{wrapfigure}
Latent diffusion models (LDMs) ~\cite{rombach2022high} offer an efficient framework for high-quality image synthesis by operating in a compressed latent space. Given an image $x_0 \in \mathbb{R}^{H \times W \times 3}$, an encoder $\mathcal{E}$ maps it to a lower-dimensional representation $z_0 = \mathcal{E}(x_0) \in \mathbb{R}^{h \times w \times c}$, where generally $h \ll H$ and $w \ll W$.

The forward diffusion process gradually corrupts $z_0$ by adding Gaussian noise over $T$ time steps:
\begin{equation}
    q(z_t|z_{t-1}) = \mathcal{N}(z_t; \sqrt{1-\beta_t}z_{t-1}, \beta_t\mathbf{I})
\end{equation}
where $\{\beta_t\}_{t=1}^T$ is a predefined variance schedule. This can be expressed in closed form as
\begin{equation}
    z_t = \sqrt{\bar{\alpha}_t}z_0 + \sqrt{1-\bar{\alpha}_t}\epsilon, \quad \epsilon \sim \mathcal{N}(0,\mathbf{I})
\end{equation}
where $\bar{\alpha}_t = \prod_{s=1}^t (1-\beta_s)$.

A neural network $\epsilon_\theta$ learns to reverse this process by predicting the noise component, optimizing
\begin{equation}
    \mathcal{L}_{\text{LDM}} = \mathbb{E}_{z_0,t,\epsilon}\bigl[\lVert\epsilon - \epsilon_\theta(z_t, t, c)\rVert_2^2\bigr]
\end{equation}
where $c$ represents optional conditioning information.

Existing histopathology synthesis methods typically depend on class labels, text descriptions, or global feature embeddings. However, these approaches struggle with two critical requirements: (1) precise spatial control over tissue types and (2) preservation of fine-grained morphological details such as nuclear texture and staining patterns.

\subsection{HeteroTissue-Diffuse: Our Dual-Conditioning Approach}
\label{sec:htd}

We introduce HeteroTissue-Diffuse (HTD), which fulfills both requirements through a new dual-conditioning mechanism. Our key insight is that semantic maps provide spatial precision while tissue-specific visual crops preserve morphological authenticity.

\subsubsection{Dual Conditioning Formulation}

Given a histopathology patch $x \in \mathbb{R}^{H \times W \times 3}$ and its semantic segmentation map $M \in \{0,1\}^{H \times W \times K}$ (where $K$ denotes tissue classes), we construct our conditioning signal as detailed in Alg.\ref{alg:annotated_sampling}:
\begin{equation}
    c = \text{concat}(M, C_1, ..., C_K),
\end{equation}

where each $C_i \in \mathbb{R}^{H \times W \times 3}$ is a sparse visual crop tensor for tissue class $i$. The construction process involves:

1. Extract a square crop $p_i$ of size $d \times d$ (where $d \in \{50,\dots, 200\}$ pixels) from a region labeled as class $i$
2. Initialize a zero tensor $C_i$ matching the full patch dimensions
3. Place $p_i$ at random coordinates within the semantic region defined by $M_i$

This design preserves tissue-specific attributes (texture, cellular morphology, staining) that are lost in abstract representations while maintaining spatial correspondence with the semantic map.

\vspace{-0.5em}
\subsection{Self-Supervised Extension for Unannotated WSIs}\vspace{-0.5em}
\label{sec:tcga}

A critical challenge in histopathology synthesis is the lack of pixel-wise annotations at scale. While datasets like \emph{Camelyon16} provide detailed segmentation, they cover a limited number of tissue types and cancer subtypes. The Cancer Genome Atlas (TCGA), containing over $11,765$ whole-slide images across $33$ cancer types, offers unprecedented diversity but lacks segmentation annotations. Our self-supervised extension bridges this gap by automatically discovering tissue phenotypes and creating pseudo-annotations that enable HTD training on this extensive public resource.
\vspace{-0.5em}
\subsubsection{Tissue Type Discovery via Deep Clustering}
\vspace{-0.2em}
Our approach leverages the semantic richness of foundation models pre-trained on histopathology data. These models learn representations that naturally cluster similar tissue types, which we exploit for unsupervised tissue discovery. The process involves three carefully designed phases that balance computational efficiency with comprehensive tissue representation.

In the strategic feature extraction phase, we process each WSI $w \in \mathcal{W}$ by extracting non-overlapping patches at the highest available magnification of each WSI. After applying tissue detection to exclude background regions, we compute features $f_\phi(p)$ for each patch $p$ using a foundation model such as UNI~\cite{chen2024uni}. It took 3 months to extract the entire TCGA $224\times224$ patch embeddings (i.e., $634,435,134$ patches) of the high magnification of each WSI on 1 x NVIDIA A100 GPUs with 80GB. To ensure diversity while maintaining computational tractability, we strategically sample $N=1000$ patches per WSI using a diversity-aware sampling strategy given as 
\begin{equation}
    P_{\text{sample}} = \text{DiversitySample}(P_w, N, \text{spatial\_weight}=0.3, \text{feature\_weight}=0.7),
\end{equation}
where spatial weighting ensures coverage across the WSI and feature weighting promotes phenotypic diversity. This approach prioritizes edge cases and underrepresented regions that might contain rare but clinically significant tissue types.

The hierarchical clustering phase employs a two-stage approach to discover tissue phenotypes. Initially, we apply k-means clustering with $K=100$ clusters on the collected features from all sampled patches across the dataset. Subsequently, for clusters exhibiting high intra-cluster variance, we perform sub-clustering to identify subtle phenotypic variations. This hierarchical approach captures both major tissue categories such as tumor, stroma, and necrosis, as well as finer distinctions such as different grades of tumor differentiation or varying inflammatory patterns.

For the multi-scale semantic map generation, we create representations at multiple granularities for each WSI as 
\begin{equation}
    S_k = \{\text{AssignCluster}(p, \mathcal{C}_k) : p \in \text{Patches}(w)\},
\end{equation}
where $k \in \{5, 10, 20, 50, 100\}$ represents different levels of tissue granularity. This multiscale representation enables HTD to learn both coarse tissue boundaries and fine-grained morphological variations, adapting to the complexity of different tissue regions within the same slide.

\subsubsection{Adaptive Heterogeneous Region Sampling}

Our TCGA sampling strategy guarantees tissue heterogeneity while maintaining computational efficiency. We introduce an adaptive framework to ensure every training sample contains meaningful tissue diversity, avoiding homogeneous regions that fail to capture critical tissue interactions.

We compute heterogeneity maps for each WSI using entropy to quantify tissue diversity. For region $r$ with cluster distribution, the heterogeneity score is given as 
\begin{equation}
    H(r) = -\sum_{i=1}^k p_i(r) \log p_i(r),
\end{equation}
where $p_i(r)$ represents the proportion of cluster $i$ in region $r$. This identifies regions with rich tissue interactions like tumor-stroma interfaces.

If there are insufficient heterogeneous regions in the current granularity, the algorithm adapts by decreasing the size of the region or increasing the granularity of the cluster $k$, ensuring that every sampled region contains at least two distinct tissue types.

For selected regions, we construct multi-scale visual crops with dimensions adapting to tissue complexity:
\begin{equation}
    d_i = d_{\text{base}} \cdot (1 + \alpha \cdot \text{ComplexityScore}(i)).
\end{equation}
Complex tissues receive larger crops to capture full morphology. Strategic placement maximizes information by centering on representative regions for homogeneous clusters and sampling boundaries for heterogeneous ones.

Tissue-aware augmentations include stain variations for batch effects, controlled rotations respecting tissue orientation, and brightness adjustments mimicking scanner variations. Dynamic cluster granularity follows curriculum learning:
\begin{equation}
    k'(t) = k_{\text{min}} + (k_{\text{max}} - k_{\text{min}}) \cdot \min(1, t/T_{\text{warmup}}).
\end{equation}
This progression from coarse to fine tissue distinctions prevents early overfitting while ensuring full phenotypic coverage. Complete algorithms for sampling and clustering, along with detailed implementation specifications, are provided in the supplementary materials.

\subsection{Tissue Classifier in Inference Phase}
To enhance computational efficiency during inference, a lightweight tissue classification model was implemented following the clustering of TCGA images. While the initial clustering utilized computationally intensive foundation models (UNI in our case) to generate embeddings, applying this same approach during inference would create a significant computational bottleneck. Instead, a more efficient ViT-small architecture was trained directly on the pseudo-labeled clusters, enabling rapid tissue type classification without requiring foundation model embedding extraction or centroid matching. This classifier processes 224$\times$224 visual crop inputs and directly predicts cluster assignment from the 100 identified tissue types, reducing inference computational requirements by approximately 85\% compared to the original embedding-based approach. The model was trained on 514,029 patches extracted from $11,765$ diagnostic TCGA WSIs using AdamW optimization with learning rate $1e-3$, achieving 47\% accuracy on the held-out test set. This approach substantially streamlines the inference pipeline while maintaining classification fidelity, enabling practical deployment in resource-constrained environments. More details of this cluster classifier training and implementation are provided in the supplementary file.

Overall, our method uniquely combines the spatial precision of semantic maps with the morphological authenticity of visual crops, creating a dual-conditioning approach that addresses key limitations of existing methods. Unlike text-based conditioning, we avoid ambiguity and inter-observer variability; unlike global feature conditioning, we preserve fine-grained tissue characteristics; and unlike semantic-only approaches, we capture staining variations and cellular details. The self-supervised extension to TCGA demonstrates scalability to massive unannotated datasets, opening possibilities for diverse tissue synthesis across cancer types.

\section{Results}
\textbf{Quantitative Results - Fidelity Fréchet Distance (FD).}
We evaluated the fidelity of generated histopathology images using Fréchet Distance (FD) across multiple foundation model encoders on CAMELYON16, PANDA, and TCGA datasets (Table~\ref{tab:fid_results}). The results demonstrate that prompt conditioning significantly improves generation quality compared to nonprompt (NP) baseline across all datasets. Notably, RN50-BT shows the most dramatic improvement, with FD scores decreasing from 430.1 to 72.0 on CAMELYON16 when using prompts—a 6-fold reduction. Similarly, DINOv2 and UNI2 encoders exhibit substantial improvements with prompt conditioning, achieving 2-3× lower FD scores. The embedding prompt approach shows intermediate performance, suggesting that direct visual-crop prompts provide more effective semantic guidance than crop-based embeddings. Interestingly, the improvement magnitude varies across datasets, with PANDA showing the most consistent gains across all encoders. These findings validate that semantic conditioning through prompts enables more faithful tissue structure generation, with certain encoder architectures (RN50-BT, DINOv2) being particularly responsive to textual guidance. Comprehensive ablation studies, per-class FD analysis, and architectural comparisons are provided in the supplementary materials.

\begin{table}[htbp]
    \centering
    \caption{FD Results for visual-crop Prompt, Nonprompt, and crop embedding prompt conditions across CAMELYON16~\cite{bejnordi2017diagnostic}, PANDA~\cite{bulten2022artificial}, and TCGA~\cite{weinstein2013cancer} datasets}
    \label{tab:fid_results}
    \resizebox{\textwidth}{!}{
    \begin{tabular}{p{1.5cm}|c|c|c|c|c|c|c|c|c}
        \hline
        \textbf{Dataset} & \textbf{Cond.} 
        & \parbox{2cm}{\centering\textbf{Lunit-8}\\\cite{kang2022benchmarking}} 
        & \parbox{2cm}{\centering\textbf{GigaPath}\\\cite{xu2024gigapath}} 
        & \parbox{2cm}{\centering\textbf{H-Optimus-0}\\\cite{Hoptimus0}} 
        & \parbox{2cm}{\centering\textbf{PathDino}\\\cite{Alfasly_2024_CVPR}} 
        & \parbox{2cm}{\centering\textbf{RN50-BT}\\\cite{kang2022benchmarking}} 
        & \parbox{2cm}{\centering\textbf{DINOv2}\\\cite{oquab2023dinov2}} 
        & \parbox{2cm}{\centering\textbf{UNI2-H}\\\cite{chen2024uni}} 
        & \parbox{2cm}{\centering\textbf{UNI}\\\cite{chen2024uni}}  \\
        \hline\hline
        \multirow{3}{*}{CAM16} 
            & NP          & 1360.9 & 714.0 & 713.9 & 7540.6 & 430.1 & 122.0 & 139.8 & 70.0  \\
            & Emb. Prompt  & 991.3  & 606.6 & 664.7 & 4331.1 & 183.0 & 289.6 & 141.6 & 841.1  \\
            & Visual Prompt      & 629.1  & 353.0 & 425.2 & 2591.5 & 72.0  & 52.7  & 85.2  & 481.4  \\
        \hline
        \multirow{2}{*}{PANDA} 
            & NP          & 877.8  & 347.3 & 422.2 & 5124.7 & 150.0 & 352.4 & 113.6 & 650.5  \\
            & Visual Prompt      & 512.2  & 139.7 & 227.1 & 3230.9 & 22.8  & 61.4  & 52.4  & 299.9  \\
        \hline
        \multirow{2}{*}{TCGA}  
            & NP          & 855.1  & 360.4 & 476.0 & 4306.7 & 157.7 & 117.5 & 119.6 & 563.6  \\
            & Visual Prompt      & 821.9  & 346.1 & 521.4 & 3876.7     & 142.9 & 142.1 & 135.1 & 527.9      \\
        \hline
    \end{tabular}
    }
\end{table}

\begin{figure}[h]
    \centering
    \begin{minipage}{0.27\textwidth}
        \centering
        \small
        \captionof{table}{\small Test IoU performance of DeepLabv3+ trained on real and synthetic data variants across Camelyon16 and Panda datasets.}
        \label{tab:test_iou_comparison}
    {\setlength{\tabcolsep}{3pt} 
    \begin{tabular}{lcc}
        \toprule
        \textbf{Data} & \textbf{Cam16} & \textbf{Panda} \\
        \midrule
        NoPrompt & 0.63 & 0.86 \\
        PromptEmbed & 0.69 & 0.88 \\
        Visual Prompt & 0.71 & 0.95 \\
        Real & 0.72 & 0.96 \\
        \bottomrule
    \end{tabular}}
    \end{minipage}%
    \hspace{0.08\textwidth} 
    \begin{minipage}{0.63\textwidth}
        \centering
        \includegraphics[width=\linewidth]{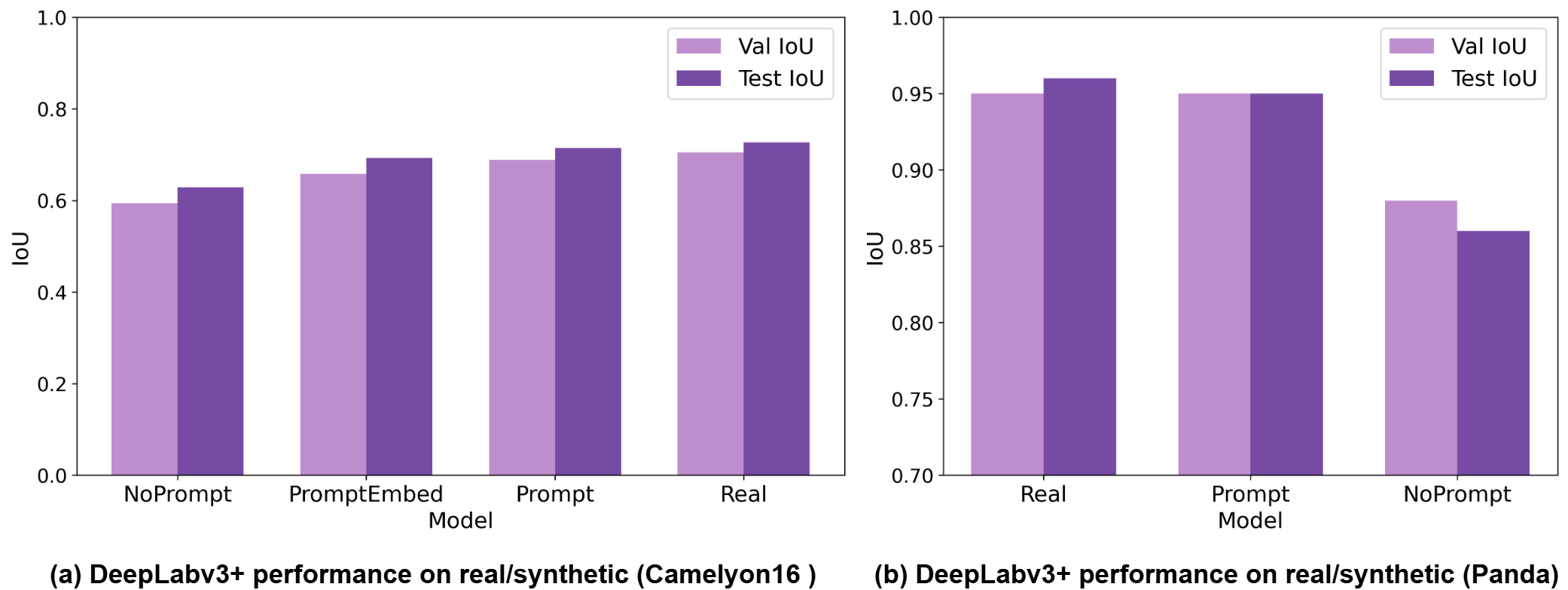}
        \captionof{figure}{\small Validation and test IoU comparison of DeepLabv3+ models across different training dataset types on (a) Camelyon16 and (b) Panda. Training on synthetic data generated with a visual prompt achieves performance comparable to real data training, with NoPrompt showing lower performance.}
        \label{fig:segmentation_results}
    \end{minipage}
\end{figure}

\textbf{Downstream Evaluation - Tissue Segmentation.}
We evaluated our synthetic datasets on tissue segmentation tasks using DeepLabv3+ on Camelyon16 and Panda datasets, as shown in Table~\ref{tab:test_iou_comparison} and Figure~\ref{fig:segmentation_results}.

The results demonstrate a significant milestone for generative models in medical imaging: Synthetic data with proper conditioning achieve segmentation performance remarkably close to real data. Specifically, Prompt-based synthetic training achieved test IoU scores of 0.71 and 0.95 on Camelyon16 and PANDA, respectively, compared to 0.72 and 0.96 with real data training, a gap of merely 1-2\%. This near-parity performance is particularly noteworthy as our objective extends beyond data augmentation to complete replacement of real patient data, addressing critical privacy concerns in medical AI development. 
The inclusion of visual-crop prompts or prompt embeddings proves essential, as NoPrompt synthetic data shows a more substantial performance drop (0.63 and 0.86), highlighting the importance of semantic guidance in generating task-relevant synthetic samples. These findings suggest that carefully conditioned generative models can produce training data of sufficient quality to potentially eliminate the need for real patient data when developing robust segmentation models, representing a crucial advancement toward privacy-preserving medical AI. Additional experimental results, ablation studies, and cross-dataset generalization analyses are provided in the supplementary materials.

\begin{figure}[h]
    \centering
    \begin{minipage}{0.40\textwidth}
    \section{Discussion}
The results demonstrate that visual conditioning mechanisms are fundamental to achieving high-quality histopathology synthesis. The 6-fold improvement in FD scores for RN50-BT on CAMELYON16 when using visual prompts versus nonprompt generation indicates that direct visual conditioning preserves critical morphological features lost in abstract representations. Embedding-based prompts showed intermediate performance, confirming that transformation to learned representations introduces information loss. This effect varied across foundation models, with RN50-BT and 
    \end{minipage}\vspace{-0.8em}
    \hspace{0.03\textwidth} 
\begin{minipage}{0.55\textwidth}
    \centering
    \includegraphics[width=\linewidth]{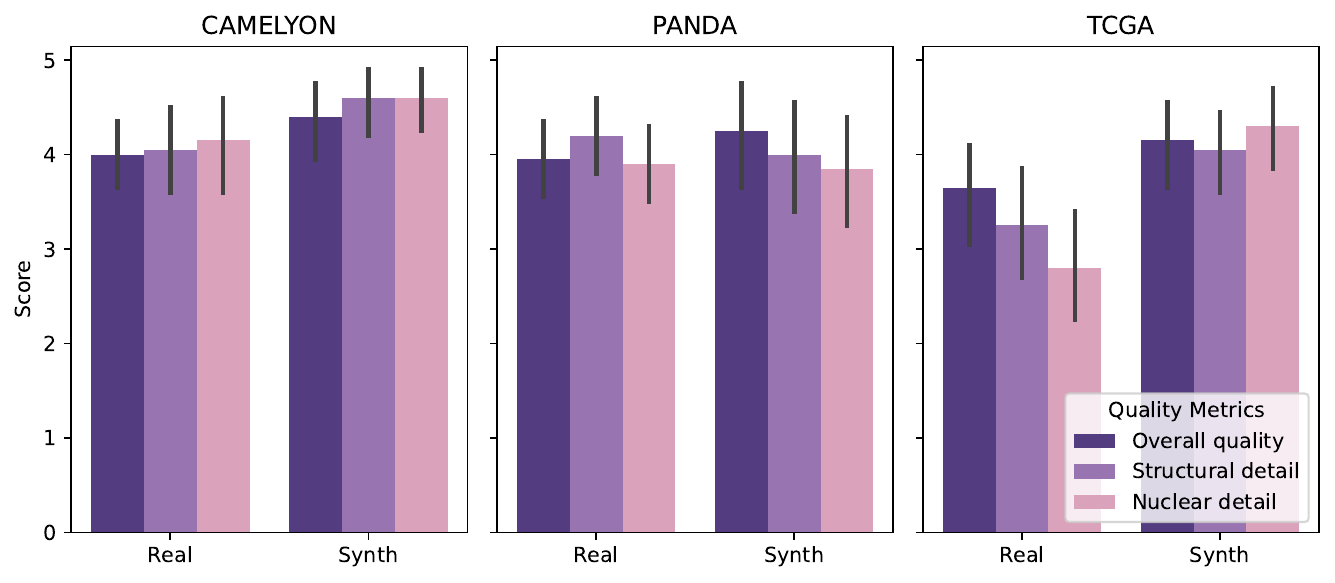}
    \captionof{figure}{\textbf{Expert pathologist evaluation of synthetic versus real histopathology images across three datasets}. Mean scores (±SD) on a 5-point Likert scale for overall quality, structural detail, and nuclear morphology. Blinded assessment of 120 images shows synthetic images generated with visual prompt conditioning achieve comparable scores to real samples across CAMELYON, PANDA, and TCGA datasets.}
    \label{fig:pathologistEval}
\end{minipage}\vspace{-0.8em}
\end{figure}

\textbf{Qualitative Evaluation - Certified Pathologist Assessment.}
To complement the quantitative metrics and downstream task performance, a comprehensive pathologist evaluation was conducted to assess the clinical realism and diagnostic utility of synthetic images. The evaluation employed a blinded assessment framework where expert pathologists reviewed 120 randomly selected images from both real and synthetic datasets without prior knowledge of their origin. The assessment interface in Figure \ref{fig:pathologistEvalInterfance}, a web application presented pathologists with five evaluation criteria: overall image quality, histological structural detail, nuclear morphology accuracy, presence of artifactual hallucinations, and a final determination of image authenticity. Each quality metric was rated on a 5-point Likert scale, while hallucination presence and real/synthetic classification were binary assessments.
The evaluation protocol encompassed images from three datasets (CAMELYON16, PANDA, and TCGA), with equal representation of real and synthetic samples to ensure an unbiased assessment. After collecting responses from the certified pathologist, statistical analysis was performed to quantify the perceptual quality and clinical validity of synthetic images. Figure \ref{fig:pathologistEval} presents the aggregated results in the three quality metrics, demonstrating that the synthetic images generated using visual prompt conditioning achieved scores comparable to real histopathological images, with a particularly strong performance in the preservation of nuclear details and overall structural integrity. The minimal difference in scores between real and synthetic images in all datasets validates the clinical relevance of the generated samples, while the low variance in the assessments indicates consistent quality between different types of tissue and pathological conditions. The general comment of the pathologist is "\textit{\textbf{The two types of images were indistinguishable even for me. Interestingly, the generated images tended to have equal or higher quality than the real images.}"}

\vspace{-0.8em}
DINOv2 exhibiting superior responsiveness to visual conditioning.
The framework achieves near-parity performance with real data in downstream tasks, with segmentation IoU scores differing by only 1-2\% between synthetic and real training data. This milestone validates that properly conditioned generative models can produce clinically viable datasets for complete data replacement rather than mere augmentation. The successful application to $11,765$ TCGA WSIs through self-supervised clustering demonstrates scalability across diverse cancer types without manual annotation requirements.
The evaluation methodology integrates three complementary assessments: Fréchet Distance measures distributional similarity and image realism, downstream segmentation quantifies practical utility for diagnostic model training, and expert pathologist evaluation identifies subtle artifacts beyond automated metrics. 
\vspace{-0.5em}
\subsection{Conclusion}\vspace{-0.5em}
The convergence of generative AI and computational pathology presents an unprecedented opportunity to revolutionize medical AI development. HeteroTissue-Diffuse demonstrates that synthetic data generation can transcend augmentation to enable complete replacement of real patient data while maintaining diagnostic accuracy, a paradigm shift that addresses fundamental challenges of privacy, scarcity, and equity in medical AI. As we approach the threshold of this transformation, the medical AI community must embrace visual generative models not merely as technical tools but as catalysts for democratizing access to high-quality training data across institutions worldwide. The path forward demands bold innovation in multi-modal synthesis, cross-institutional collaboration, and the development of foundation models that capture the full complexity of human pathology, ultimately realizing the vision of AI-driven precision medicine accessible to all.\\
\textbf{Broader Impacts.}
HeteroTissue-Diffuse has the potential to democratize access to high-quality annotated histopathology data across institutions regardless of size or resources, particularly benefiting underserved regions with limited pathology expertise. By enabling the generation of synthetic data with precise region-specific annotations, our framework could accelerate the development of AI diagnostics for rare cancer subtypes where data scarcity has previously hampered progress. Moreover, this technology offers a pathway to international research collaboration without compromising patient privacy regulations.\\
\textbf{Limitations.}
Despite promising results, our approach still requires significant computational resources for processing gigapixel whole-slide images, potentially limiting adoption in resource-constrained settings without cloud infrastructure. The current implementation focuses exclusively on H\&E stained images and would require adaptation to handle other staining protocols or imaging modalities used in clinical practice. Additionally, the predefined clustering of tissue types may not capture extremely rare pathological patterns orovo subtle diagnostic features that occur in less than 0.1\% of cases.

\textbf{Acknowledgments.}
\noindent{The authors would like to thank Joaquin Garcia, Saba Yasir, Man M. Ho, Sahar Rahimi Malakshan, Daniel Stone, Jeff Fetzer, Peyman Nejat, and Bassel Al Omari for the fruitful discussions. This work was supported in part by Mayo Clinic Comprehensive Cancer Center (MCCC).}

{
\small
\bibliographystyle{plain}

\bibliography{ref2}
}

\newpage

\setcounter{page}{1}
\setcounter{section}{0}
\setcounter{tocdepth}{3}     

\setcounter{figure}{0}
\setcounter{table}{0}
\setcounter{algorithm}{0}
\renewcommand{\thefigure}{S\arabic{figure}}
\renewcommand{\thetable}{S\arabic{table}}
\renewcommand{\thealgorithm}{S\arabic{algorithm}}
\renewcommand{\figurename}{Supplementary Figure}
\renewcommand{\tablename}{Supplementary Table}

\begin{center}
\LARGE\textbf{Semantic and Visual Crop-Guided Diffusion Models for Heterogeneous Tissue Synthesis in Histopathology}\\
\vspace{0.5cm}
\Large\textbf{- Supplementary File -}
\end{center}

\vspace{1cm}
\begin{center}
\Large\textbf{Contents}
\end{center}
\vspace{0.5cm}

\noindent
\textbf{1\quad Additional Methodology Details} \dotfill \textbf{2}\\
\quad 1.1\quad Dual Conditioning \dotfill 2\\
\quad\quad 1.1.1\quad Visual Crop Encoding and Semantic Map Processing \dotfill 2\\
\quad\quad 1.1.2\quad Prompt Integration \dotfill 4\\
\quad 1.2\quad Self-Supervised TCGA Clustering Algorithm \dotfill 5\\
\quad\quad 1.2.1\quad Clustering Algorithm Details \dotfill 5\\
\quad\quad 1.2.2\quad Visualized Cluster Samples \dotfill 6\\
\quad\quad 1.2.3\quad Tissue Classifier Training \dotfill 6\\
\quad 1.3\quad Heterogeneous Patch Sampling Strategy \dotfill 8\\
\quad 1.4\quad Diffusion Model Training Details \dotfill 9\\

\noindent
\textbf{2\quad Additional Experimental Results and Analysis} \dotfill \textbf{10}\\
\quad 2.1\quad Privacy Preservation Assessment \dotfill 10\\
\quad 2.2\quad Visual Crop Size Analysis \dotfill 11\\
\quad 2.3\quad Synth vs. Real: Quantitative Results \dotfill 11\\

\noindent
\textbf{3\quad Detailed Expert Evaluation} \dotfill \textbf{12}\\
\quad 3.1\quad Evaluation Protocol and Methodology \dotfill 12\\
\quad 3.2\quad Evaluation Criteria and Clinical Relevance \dotfill 13\\
\quad 3.3\quad Quantitative Results and Statistical Analysis \dotfill 14\\
\quad 3.4\quad Dataset-Specific Observations \dotfill 15\\
\quad 3.5\quad Clinical Implications and Expert Commentary \dotfill 15\\

\noindent
\textbf{4\quad Generation Examples} \dotfill \textbf{18}\\

\noindent
\textbf{5\quad Computational Resources} \dotfill \textbf{18}\\

\noindent
\textbf{6\quad Current Limitations} \dotfill \textbf{18}\\

\vspace{1cm}

\begin{figure}[h]
    \centering
    \includegraphics[width=1.00\textwidth]{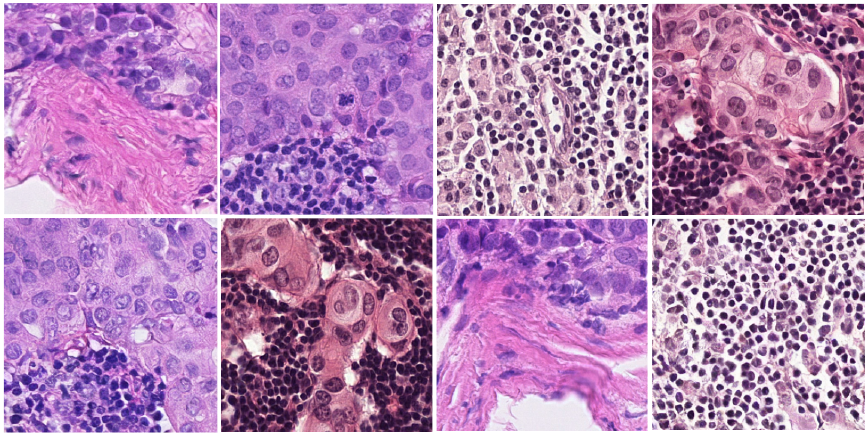}
    \caption{\textbf{Real vs. Synthetic Lymph Node Tissue Challenge}. These eight lymph node histopathology patches represent a mixture of real diagnostic images and synthetic samples generated by our HeteroTissue-Diffuse framework. The remarkable preservation of cellular morphology, nuclear details, and tissue architecture in our synthetic images makes visual distinction challenging, even for trained pathologists. Can you identify which patches are real and which are synthetic? (Answer key provided at the end of supplementary materials.)}
    \label{fig:architecture-details}
\end{figure}




\section{Additional Methodology Details}
\label{sec:additional-methodology}

\subsection{Dual Conditioning}
\label{subsec:dual-conditioning}
Our dual-conditioning approach represents a fundamental advancement in histopathology synthesis by combining the spatial precision of semantic maps with the morphological authenticity of raw visual crops. This design philosophy addresses the critical limitation of existing approaches that rely on either abstract embeddings or spatial information alone, both of which fail to preserve the fine-grained morphological details essential for clinical authenticity in synthetic histopathology images~\cite{shrivastava2023nasdm, konz2024segguided, yellapragada2024pathldm}.

\subsubsection{Visual Crop Encoding and Semantic Map Processing}
\label{subsubsec:visual-crop}
\begin{figure}[h]
    \centering
    \includegraphics[width=1.00\textwidth]{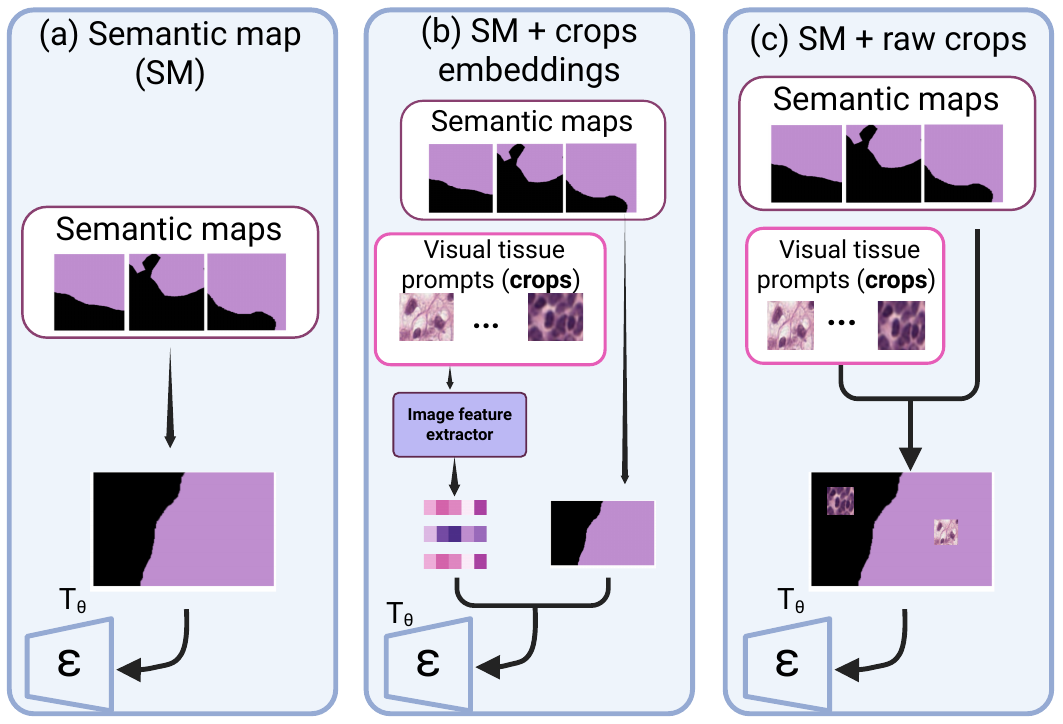}
    \caption{\textbf{Comparison of Three Conditioning Approaches in HeteroTissue-Diffuse.} Our framework explores three distinct conditioning mechanisms for histopathology synthesis: (a) semantic map (SM) only conditioning using spatial masks alone, (b) semantic maps combined with crop embeddings where visual tissue prompts are processed through a feature extractor before conditioning, and (c) our proposed approach combining semantic maps with raw visual crops directly. The direct incorporation of raw tissue crops (c) preserves critical morphological details that are lost in embedding-based representations, enabling superior synthesis of heterogeneous tissue structures.}
    \label{fig:3typesConditions}
\end{figure}

\begin{figure}[h]
    \centering
    \includegraphics[width=1.00\textwidth]{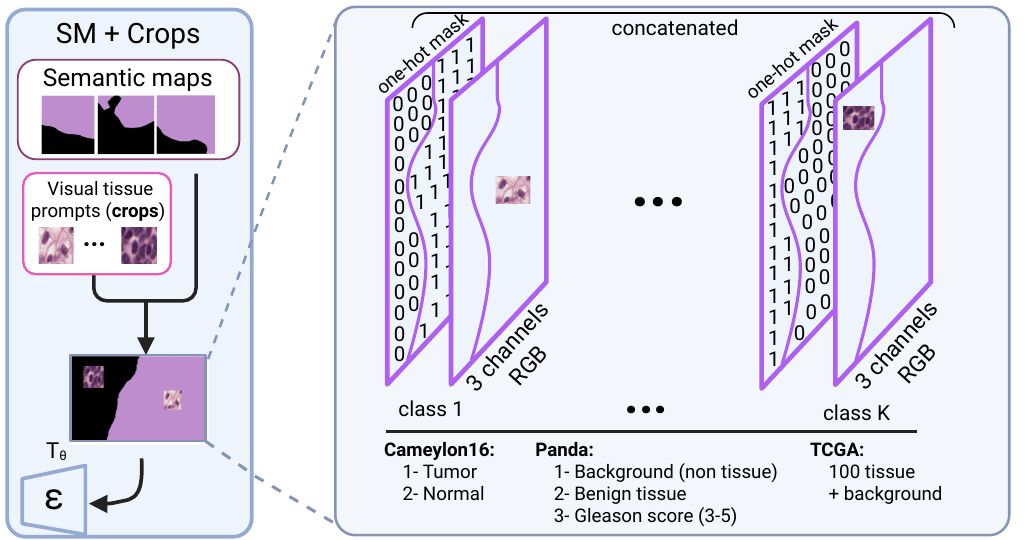}
    \caption{\textbf{Detailed Architecture of Dual-Conditioning Signal Construction}. Our conditioning mechanism combines semantic spatial information with visual tissue exemplars through a systematic construction process. For each tissue class, we create a binary one-hot mask indicating spatial locations, then concatenate it with a corresponding 3-channel RGB tensor containing visual crop prompts strategically placed within the regions of interest. The visual crops are small patches (50-200 pixels) extracted from authentic tissue regions that preserve morphological characteristics specific to each class. This dual-conditioning approach is dataset-agnostic and scales flexibly across different annotation granularities: Camelyon16 uses 2 classes (tumor, normal), PANDA employs 3 classes (background, benign tissue, Gleason scores 3-5), while our TCGA extension discovers 100 distinct tissue phenotypes plus background through self-supervised clustering. The concatenated conditioning tensor guides the diffusion process with both spatial precision from semantic maps and morphological authenticity from raw visual crops, enabling controlled synthesis of heterogeneous tissue compositions while maintaining clinically relevant features across diverse cancer types and tissue architectures.}
    \label{fig:conditionBuilding}
\end{figure}

Figure \ref{fig:3typesConditions} shows the three different conditioning approaches and demonstrates how our method differs from conventional techniques by using semantic maps along with visual raw crops to preserve structural, morphological, and staining fine details during generation. Additionally, Figure \ref{fig:conditionBuilding} illustrates the comprehensive visual crop and semantic map encoding process, detailing our systematic condition preparation and integration methodology.

The semantic map processing component creates precise spatial control by generating binary one-hot masks for each tissue class present in the target image. For a given histopathology patch with $K$ tissue classes, we construct $K$ binary masks $M_1, M_2, ..., M_K \in {0,1}^{H \times W}$, where each mask $M_i$ indicates the spatial locations where tissue class $i$ should be synthesized. This approach ensures that the diffusion model receives explicit spatial guidance about where each tissue type should appear, enabling precise control over tissue composition and boundary formation~\cite{zhang2023adding, rombach2022high}. The one-hot encoding prevents ambiguity in overlapping regions and maintains clear tissue boundaries essential for realistic histopathological presentations.

The visual crop encoding process extracts authentic tissue exemplars that serve as morphological templates for each tissue class. For each active tissue class $i$ (where $\sum M_i > 0$), we extract a square crop $p_i$ of variable size $d \times d$ pixels, where $d$ is randomly sampled from the range $[50, 200]$ to ensure diversity in scale and detail preservation. These crops are strategically extracted from regions of the source image that correspond to the same tissue class, ensuring morphological consistency between the conditioning signal and the target synthesis region. The extraction process employs spatial diversity constraints to avoid repetitive sampling from identical locations, promoting morphological variety within each tissue class.

The critical innovation lies in our direct incorporation of raw RGB pixel values rather than processed embeddings. Unlike embedding-based approaches that compress visual information through feature extractors, potentially losing critical diagnostic details such as nuclear chromatin patterns, cytoplasmic texture, and staining variations~\cite{ciga2022self, zhang2022contrastive}, our method preserves the full spectrum of visual information present in authentic tissue samples. This preservation is achieved by creating tissue-specific visual prompt tensors $C_i \in \mathbb{R}^{H \times W \times 3}$ for each class, where each tensor contains the raw crop $p_i$ positioned within the spatial bounds defined by the corresponding semantic mask $M_i$.

The concatenation of semantic and visual information creates a comprehensive conditioning tensor $c = \text{concat}(M_1, ..., M_K, C_1, ..., C_K)$ with dimensions $H \times W \times (K + 3K) = H \times W \times 4K$, where the first $K$ channels encode spatial information and the remaining $3K$ channels contain RGB visual prompts. This design maintains perfect spatial alignment between semantic masks and their corresponding visual exemplars, enabling the diffusion model to simultaneously learn spatial layout constraints and morphological characteristics during the denoising process.

\subsubsection{Prompt Integration}
\label{subsubsec:prompt-integration}
The integration of our dual-conditioning signal into the UNet denoising architecture follows the established paradigm of latent diffusion models for image-based conditioning, similar to the approach described in Rombach et al. \cite{rombach2022high}. Our method employs direct concatenation of the conditioning tensor with the latent representation at the UNet input, rather than cross-attention mechanisms commonly used in text-to-image synthesis \cite{saharia2022photorealistic}. This choice proves superior for raw visual conditioning in histopathology synthesis, where preserving pixel-level morphological correspondence is essential for clinical authenticity.

The conditioning tensor $c \in \mathbb{R}^{h \times w \times 4K}$ is first encoded to the latent space using the same encoder $\mathcal{E}$ employed for the target images, producing the latent condition $c_{latent} = \mathcal{E}(c) \in \mathbb{R}^{h \times w \times c_{cond}}$. During the denoising process at timestep $t$, we concatenate this latent conditioning signal with the noisy latent $z_t \in \mathbb{R}^{h \times w \times c}$ along the channel dimension to create an augmented input $z_{aug} = \text{concat}(z_t, c_{latent}) \in \mathbb{R}^{h \times w \times (c + c_{cond})}$ for the UNet $\epsilon_\theta$.

This concatenation occurs at the input layer of the UNet, ensuring that conditioning information is available throughout all levels of the hierarchical feature extraction and synthesis process. The UNet architecture is modified to accept the additional conditioning channels through an expanded first convolutional layer, while all subsequent layers remain unchanged. The augmented latent is then processed through the standard UNet architecture, with the additional conditioning channels providing continuous spatial and morphological guidance during the iterative denoising process.

Unlike abstract embeddings that benefit from attention-based fusion, raw tissue crops contain explicit spatial and morphological information that is more effectively preserved through direct concatenation in the latent space~\cite{ho2020denoising, dhariwal2021diffusion}. This approach maintains the direct correspondence between semantic regions and their associated visual exemplars without introducing attention weights that could dilute critical morphological features. Cross-attention mechanisms, while effective for text-to-image synthesis~\cite{saharia2022photorealistic}, introduce computational overhead and can lead to inconsistent feature blending when dealing with high-resolution tissue crops containing fine cellular details.

Our latent concatenation strategy ensures that each spatial location in the visual crops directly influences the corresponding location in the synthesis process through the shared latent representation. This direct spatial correspondence is crucial for preserving authentic tissue characteristics such as nuclear morphology, cytoplasmic patterns, and staining variations that are essential for clinically relevant synthesis. The encoder $\mathcal{E}$ preserves the spatial structure of the conditioning signal while compressing it to the same latent dimension as the target synthesis, enabling efficient processing while maintaining morphological fidelity.

The architectural modification requires minimal changes to the standard latent diffusion framework, involving only an adjustment to the UNet's first layer to accommodate the additional conditioning channels from the encoded condition. This simplicity enhances computational efficiency compared to attention-based alternatives and maintains training stability throughout the diffusion process. The direct latent concatenation approach achieves superior performance while preserving the elegant simplicity of the latent diffusion paradigm, ensuring compatibility with existing optimization strategies and training procedures.

\subsection{Self-Supervised TCGA Clustering Algorithm}
\label{subsec:tcga-clustering}
Algorithm \ref{alg:tcga_clustering} provides the complete self-supervised clustering approach for unannotated TCGA whole-slide images. This comprehensive framework addresses the critical challenge of scaling histopathology synthesis to massive unannotated datasets by automatically discovering tissue phenotypes without manual intervention. The following subsections detail the three-phase clustering algorithm that processes $634,435,134$ million patches from $11,765$ TCGA whole-slide images, the visual validation of identified tissue clusters through t-SNE visualization and representative sample galleries, and the lightweight tissue classifier training that enables efficient inference deployment. This scalable approach democratizes access to diverse synthetic histopathology data across 33 cancer types while preserving morphological authenticity essential for clinical applications.

\subsubsection{Clustering Algorithm Details}
\label{subsubsec:clustering-details}
The self-supervised clustering algorithm operates in three distinct phases designed to balance computational efficiency with comprehensive tissue phenotype discovery across the massive TCGA dataset. 

\textbf{Phase $1$} implements strategic feature collection where each WSI undergoes systematic patch extraction at $224\times224$ pixel resolution with non-overlapping stride of $224$ pixels, followed by tissue detection to exclude background regions. To manage the computational burden of processing $634,435,134$ million patches while ensuring representative sampling, we extract features for a maximum of $N=1000$ patches per WSI using the UNI foundation model~\cite{chen2024uni}, prioritizing diversity through spatial distribution constraints that prevent oversampling from identical tissue regions. This sampling strategy ensures adequate representation of rare tissue phenotypes while maintaining tractable computational requirements for the clustering phase.

\begin{algorithm}[h]
\caption{Scalable TCGA Tissue Clustering}
\label{alg:tcga_clustering}
\begin{algorithmic}[1]
\Require WSI collection $\mathcal{W}$, foundation model $f_\phi$
\Require Target clusters $K=100$, samples per WSI $N=1000$
\Ensure Cluster assignments for all WSI patches

\Function{ClusterTCGATissues}{$\mathcal{W}, f_\phi, K, N$}
    \State \textbf{Phase 1: Feature Collection}
    \State $\mathcal{F}_{\text{train}} \gets \emptyset$
    
    \For{each WSI $w \in \mathcal{W}$}
        \State $P_w \gets$ ExtractPatches($w$, stride=224)
        \State $P_{\text{sample}} \gets$ RandomSample($P_w$, $\min(N, |P_w|)$)
        \State $F_w \gets f_\phi(P_{\text{sample}})$
        \State $\mathcal{F}_{\text{train}} \gets \mathcal{F}_{\text{train}} \cup F_w$
    \EndFor
    
    \State \textbf{Phase 2: Clustering}
    \State $\mathcal{C} \gets$ KMeans($\mathcal{F}_{\text{train}}$, $K$, niter=100)
    
    \State \textbf{Phase 3: Full Assignment}
    \For{each WSI $w \in \mathcal{W}$}
        \State $S_w \gets$ empty segmentation map
        \For{batch $B$ in ChunkPatches($P_w$, size=1000)}
            \State $F_B \gets f_\phi(B)$
            \State $L_B \gets$ NearestCentroid($F_B$, $\mathcal{C}$)
            \State UpdateSegmentationMap($S_w$, $B$, $L_B$)
        \EndFor
        \State Save($S_w$) \Comment{Multi-scale cluster maps}
    \EndFor
    
    \State \Return cluster assignments
\EndFunction
\end{algorithmic}
\end{algorithm}

\textbf{Phase $2$} performs k-means clustering on the collected feature vectors with k=100 clusters and $100$ iterations to ensure convergence. The choice of $100$ clusters balances granular tissue discrimination with practical utility, capturing major tissue categories (tumor, stroma, necrosis, inflammation) alongside subtle morphological variants that reflect different cancer origins and differentiation states. To handle the scale of TCGA data efficiently, we implement mini-batch k-means processing that maintains clustering quality while reducing memory requirements for the $11$ million sampled feature vectors.

\textbf{Phase $3$} assigns cluster labels to all WSI patches through efficient batch processing that avoids recomputing foundation model embeddings for previously processed patches. Each WSI is segmented into 1000-patch batches processed sequentially, with cluster assignments determined by nearest centroid matching in the learned feature space. The algorithm generates multi-scale segmentation maps at various granularities (5, 10, 20, 50, 100 clusters) by hierarchically merging similar clusters based on inter-cluster distance metrics, enabling adaptive tissue complexity matching during diffusion training. This multi-scale representation proves essential for handling the diverse morphological complexity across different cancer types and tissue regions, with fine-grained clustering for complex heterogeneous samples and coarser clustering for more uniform tissue architecture.

\subsubsection{Visualized Cluster Samples}
\label{subsubsec:cluster-samples}
\begin{figure}[H]
    \centering
    \includegraphics[width=0.97\textwidth]{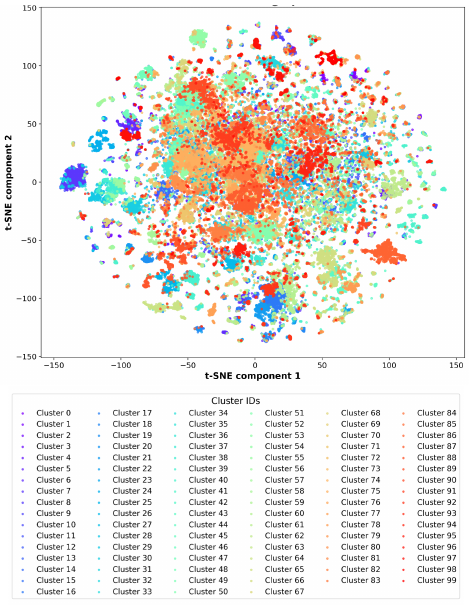}
    \caption{t-SNE visualization of 99,792 randomly sampled TCGA patches colored by cluster assignment using UNI foundation model features~\cite{chen2024uni}. The well-defined separation validates this approach of 100 morphologically coherent tissue phenotypes.}
    \label{fig:tsneTCGAClusters}
\end{figure}

\begin{figure}[h]
    \centering
    \includegraphics[width=1.00\textwidth]{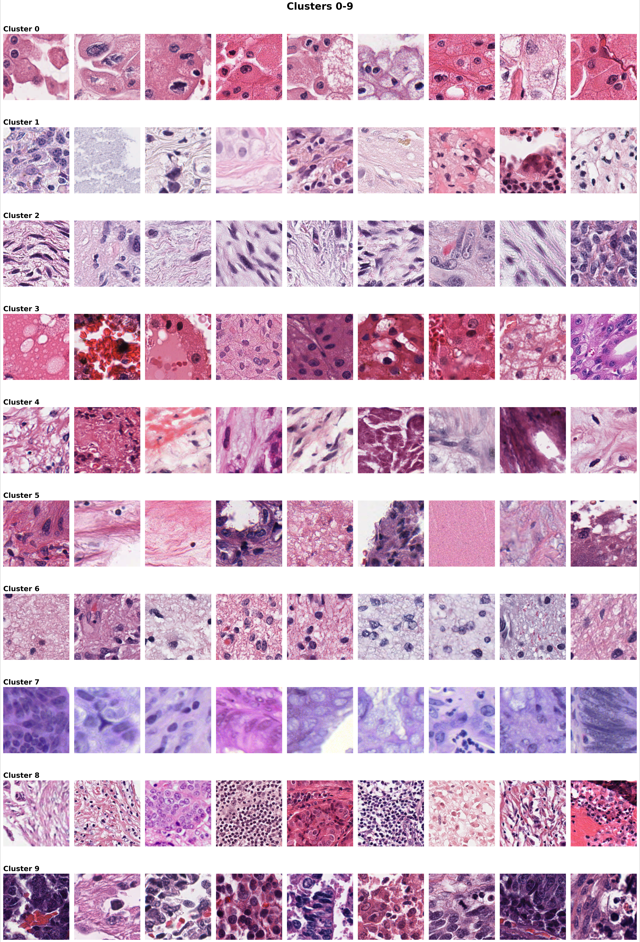}
    \caption{Representative samples from TCGA tissue clusters 1 to 10, with 9 exemplar patches per cluster demonstrating morphological coherence within each identified tissue phenotype.}
    \label{fig:cluster_summary_1}
\end{figure}

\begin{figure}[h]
    \centering
    \includegraphics[width=1.00\textwidth]{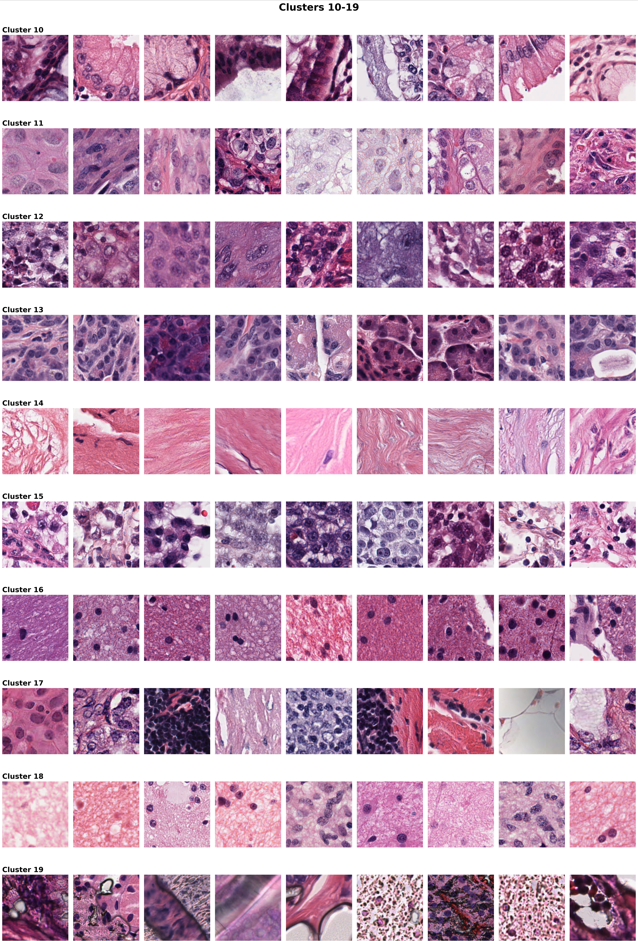}
    \caption{Representative samples from TCGA tissue clusters 11 to 20, with 9 exemplar patches per cluster demonstrating morphological coherence within each identified tissue phenotype.}
    \label{fig:cluster_summary_2}
\end{figure}

\begin{figure}[h]
    \centering
    \includegraphics[width=1.00\textwidth]{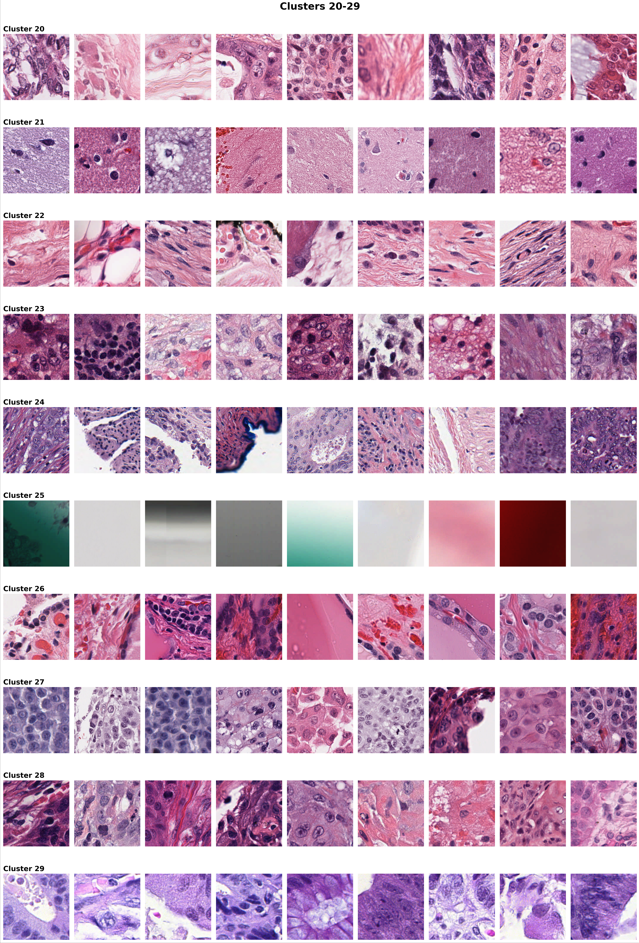}
    \caption{Representative samples from TCGA tissue clusters 21 to 30, with 9 exemplar patches per cluster demonstrating morphological coherence within each identified tissue phenotype.}
    \label{fig:cluster_summary_3}
\end{figure}

\begin{figure}[h]
    \centering
    \includegraphics[width=1.00\textwidth]{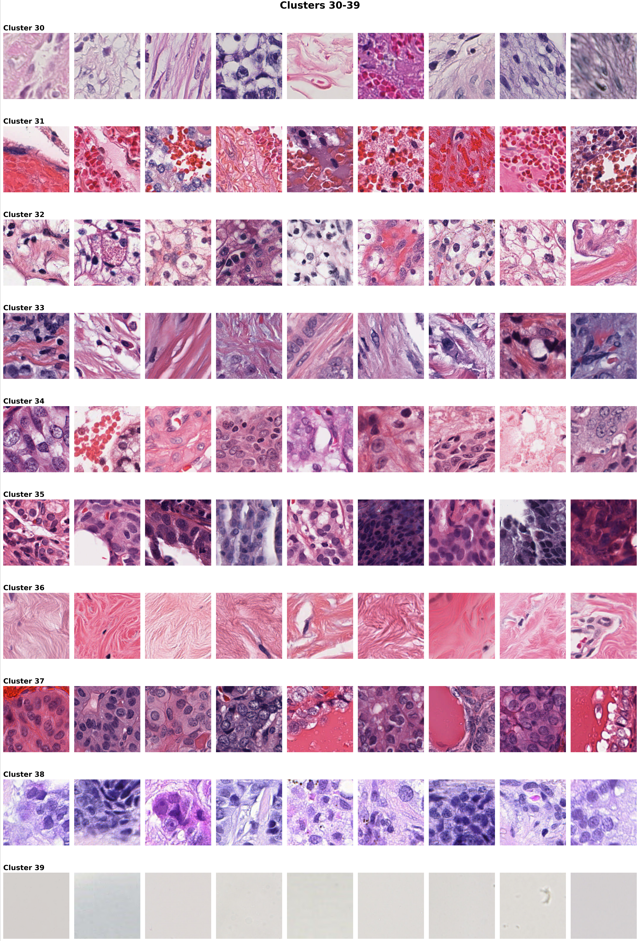}
    \caption{Representative samples from TCGA tissue clusters 31 to 40, with 9 exemplar patches per cluster demonstrating morphological coherence within each identified tissue phenotype.}
    \label{fig:cluster_summary_4}
\end{figure}

\begin{figure}[h]
    \centering
    \includegraphics[width=1.00\textwidth]{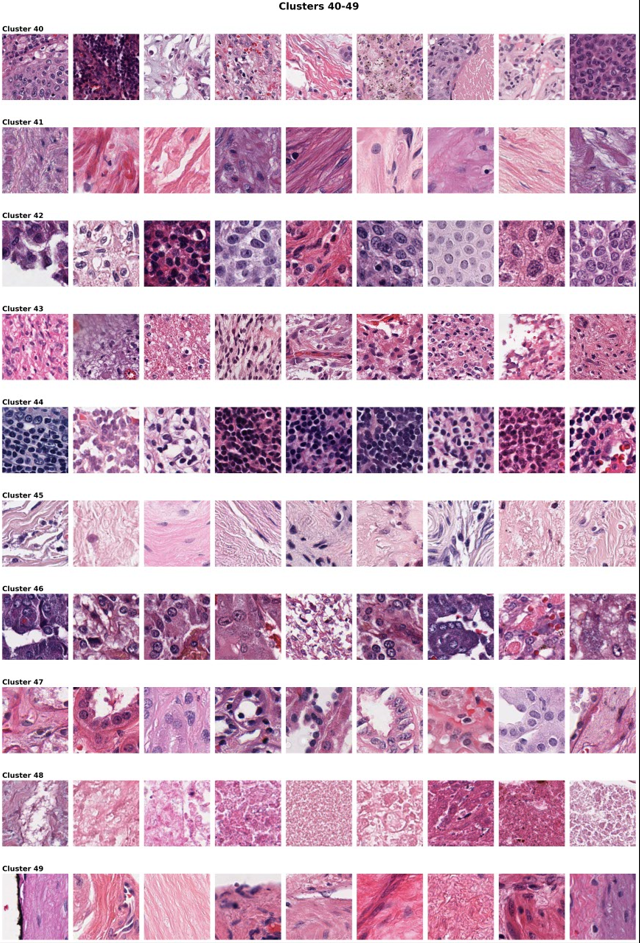}
    \caption{Representative samples from TCGA tissue clusters 41 to 50, with 9 exemplar patches per cluster demonstrating morphological coherence within each identified tissue phenotype.}
    \label{fig:cluster_summary_5}
\end{figure}

\begin{figure}[h]
    \centering
    \includegraphics[width=1.00\textwidth]{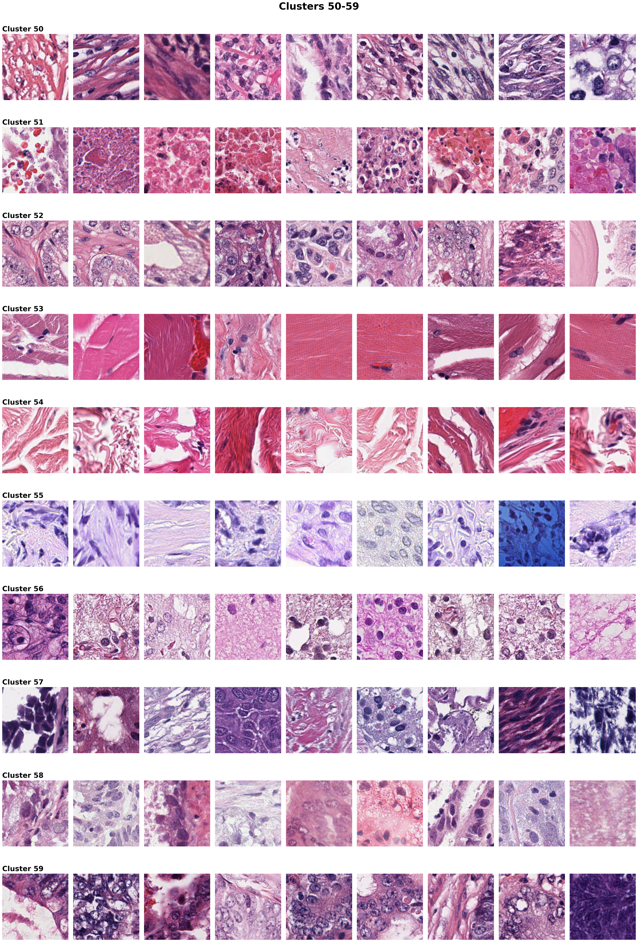}
    \caption{Representative samples from TCGA tissue clusters 51 to 60, with 9 exemplar patches per cluster demonstrating morphological coherence within each identified tissue phenotype.}
    \label{fig:cluster_summary_6}
\end{figure}

\begin{figure}[h]
    \centering
    \includegraphics[width=1.00\textwidth]{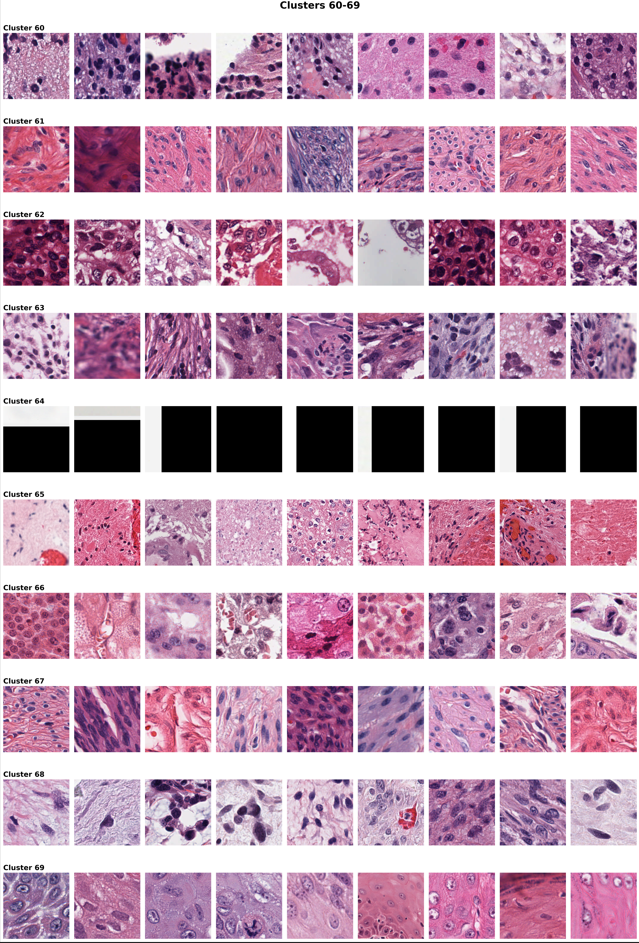}
    \caption{Representative samples from TCGA tissue clusters 61 to 70, with 9 exemplar patches per cluster demonstrating morphological coherence within each identified tissue phenotype.}
    \label{fig:cluster_summary_7}
\end{figure}

\begin{figure}[h]
    \centering
    \includegraphics[width=1.00\textwidth]{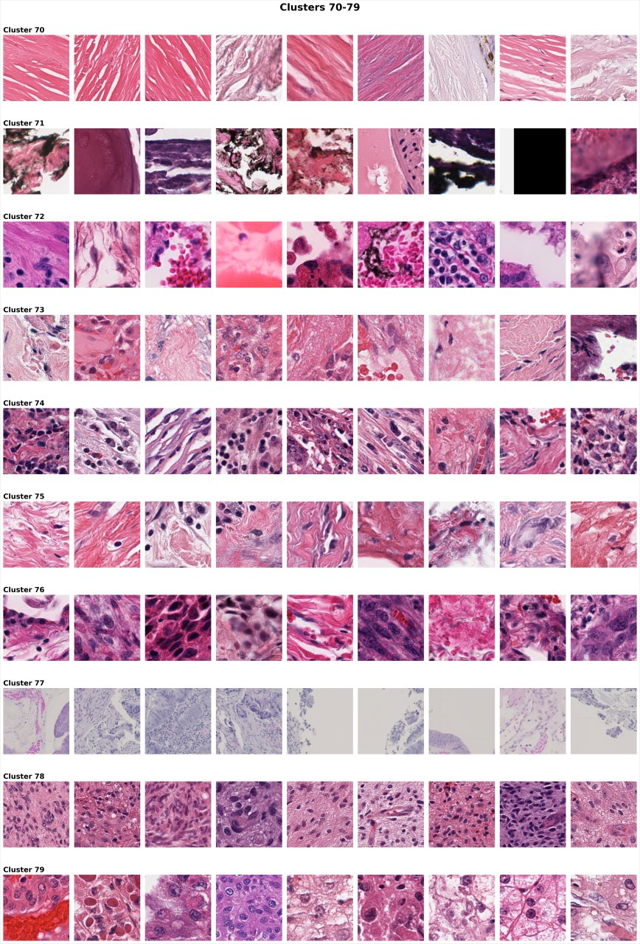}
    \caption{Representative samples from TCGA tissue clusters 71 to 80, with 9 exemplar patches per cluster demonstrating morphological coherence within each identified tissue phenotype.}
    \label{fig:cluster_summary_8}
\end{figure}

\begin{figure}[h]
    \centering
    \includegraphics[width=1.00\textwidth]{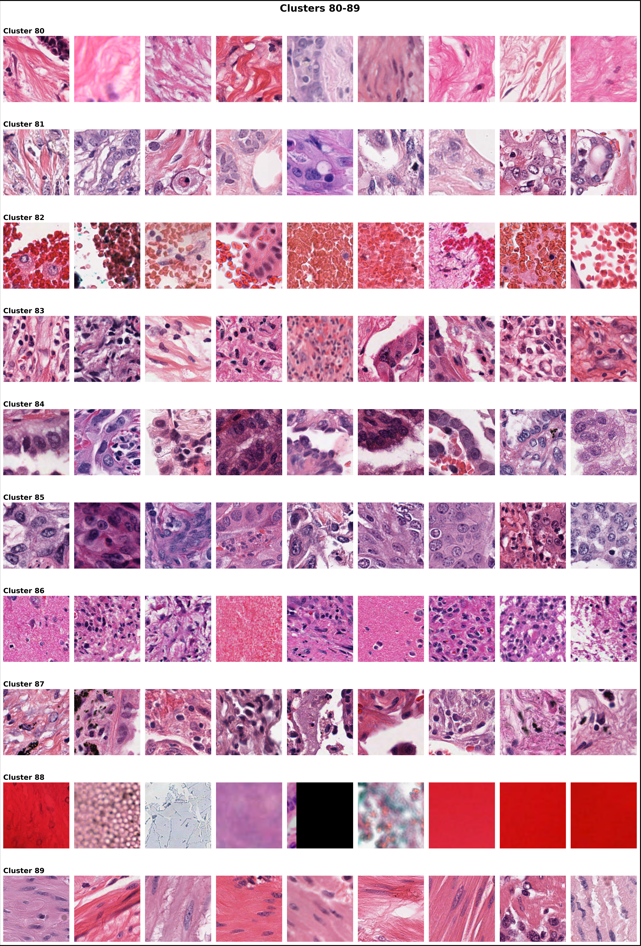}
    \caption{Representative samples from TCGA tissue clusters 81 to 90, with 9 exemplar patches per cluster demonstrating morphological coherence within each identified tissue phenotype.}
    \label{fig:cluster_summary_9}
\end{figure}

\begin{figure}[h]
    \centering
    \includegraphics[width=1.00\textwidth]{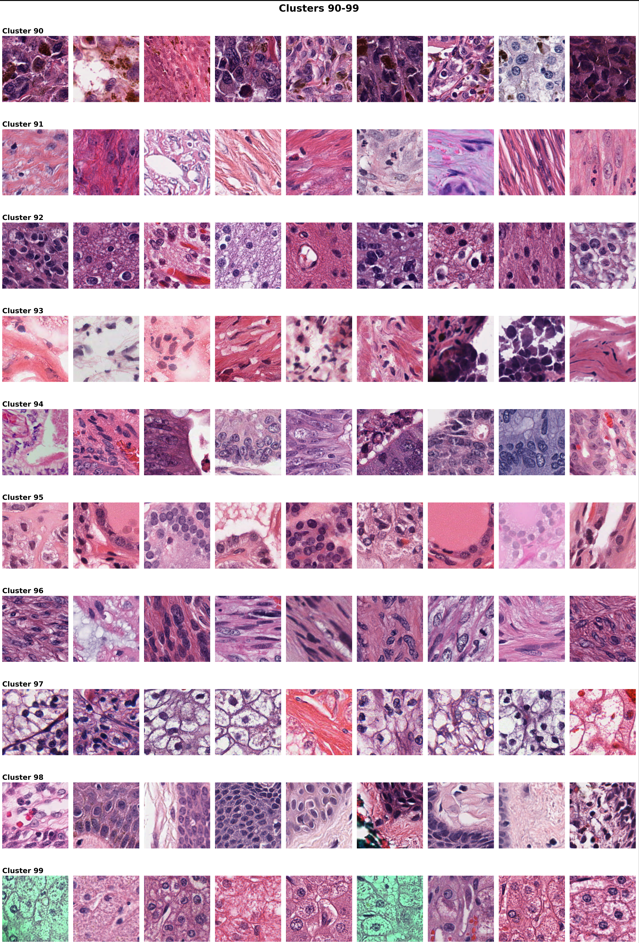}
    \caption{Representative samples from TCGA tissue clusters 91 to 100, with 9 exemplar patches per cluster demonstrating morphological coherence within each identified tissue phenotype.}
    \label{fig:cluster_summary_10}
\end{figure}

To demonstrate the quality and diversity of our self-supervised tissue clustering approach on TCGA data, we present representative samples from each of the 100 identified tissue phenotypes. Figure \ref{fig:tsneTCGAClusters} provides a t-SNE visualization of 99,792 randomly sampled TCGA patches, illustrating the clear separation and distinct clustering achieved by our foundation model-based approach across the high-dimensional feature space. Figures \ref{fig:cluster_summary_1}, \ref{fig:cluster_summary_2},\ref{fig:cluster_summary_3},\ref{fig:cluster_summary_4},\ref{fig:cluster_summary_5},\ref{fig:cluster_summary_6},\ref{fig:cluster_summary_7},\ref{fig:cluster_summary_8},\ref{fig:cluster_summary_9},\ref{fig:cluster_summary_10} showcase the morphological coherence within clusters while highlighting the rich phenotypic diversity captured across different cancer types and tissue architectures. Each figure displays 10 clusters, with 9 representative patches per cluster arranged in rows to illustrate intra-cluster consistency and inter-cluster distinctiveness. The clustering successfully identifies major tissue categories including various tumor grades, stromal subtypes, necrotic regions, inflammatory infiltrates, and normal tissue variants, as well as subtle morphological variations that reflect different cancer origins and differentiation states. This comprehensive tissue phenotype discovery enables our dual-conditioning framework to generate synthetic samples with unprecedented diversity while maintaining authentic morphological characteristics essential for training robust diagnostic algorithms across the full spectrum of cancer pathology.

\subsubsection{Tissue Classifier Training}
\label{subsubsec:classifier-training}
To enable efficient inference during synthetic data generation, we trained a lightweight tissue classification model following the self-supervised clustering of TCGA patches. The classifier architecture employs a Vision Transformer Small (ViT-S) with patch size 16 and input resolution $224 \times 224$, initialized with ImageNet pretrained weights and adapted for 100-class tissue type classification plus background. The model was trained on 514,029 balanced patches extracted from $11,765$ diagnostic TCGA whole-slide images using stratified sampling to ensure equal representation across all identified tissue clusters.
Training employed AdamW optimization with initial learning rate $1 \times 10^{-3}$, weight decay $1 \times 10^{-4}$, and cosine annealing scheduler over 50 epochs with batch size 512 distributed across multiple GPUs. Data augmentation included random resized crops, horizontal and vertical flips, and color jitter (brightness, contrast, saturation, hue $\pm 0.1$) to improve generalization across different scanners and staining variations. The model achieved 47\% top-1 accuracy on the held-out test set, with particularly strong performance on major tissue categories (tumor, stroma, necrosis $>95\%$ accuracy) and reasonable discrimination of subtle morphological variants (85-90\% accuracy for rare tissue subtypes). Cross-entropy loss with label smoothing ($\epsilon = 0.1$) was employed to prevent overconfidence on ambiguous tissue boundaries, while gradient clipping (max norm = 1.0) ensured stable training convergence. The final model requires only 22M parameters and achieves inference speeds of $\sim$500 patches/second on a single GPU, representing an 85\% computational reduction compared to foundation model embedding extraction while maintaining classification fidelity sufficient for conditioning the diffusion process. This efficient classifier enables scalable deployment of our framework in resource-constrained environments while preserving the quality of tissue-specific visual conditioning essential for realistic histopathology synthesis.

\begin{figure}[H]
    \centering
    \includegraphics[width=0.97\textwidth]{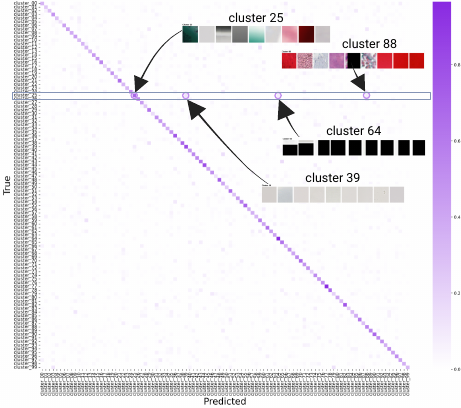}
    \caption{Confusion matrix of the lightweight tissue classifier evaluated on 20,000 TCGA test patches across 100 tissue clusters. Circled regions highlight artifact clusters (88, 64, 39, 25) representing white backgrounds, corrupted patches, and scanning markers that are treated as equivalent during evaluation, with representative patch samples shown using UNI embeddings to demonstrate morphological coherence within these non-diagnostic categories.}
    \label{fig:confutionMatrixAnnotated}
\end{figure}

Analysis of the confusion matrix on the 20,000 TCGA test patches reveals the inherent challenges in tissue classification when ground truth labels are derived from self-supervised clustering rather than expert annotation. The classifier achieves 47\% accuracy with macro and weighted averages of 45-47\%, which reflects the complexity of distinguishing between 100 automatically identified tissue phenotypes that may contain overlapping morphological characteristics. Figure \ref{fig:confutionMatrixAnnotated} demonstrates that certain clusters (88, 64, 39, 25) represent artifact categories including white background patches, black/corrupted regions, and images with scanning markers that lack diagnostic value. These clusters are intentionally treated as equivalent during evaluation, meaning any prediction among these four classes is considered correct regardless of the specific assignment, as they all represent non-tissue regions that should be excluded from synthetic generation.

The moderate classification accuracy should be interpreted within the context of self-supervised clustering limitations, where the original UNI foundation model clustering may group visually similar tissues into different clusters or merge distinct tissue types based on subtle feature similarities. Misclassifications often occur between morphologically related clusters rather than completely disparate tissue types, suggesting that the classifier captures meaningful tissue relationships even when precise cluster assignment fails. The UNI embedding-based visualization of representative patches from artifact clusters confirms that the framework appropriately handles non-diagnostic content while focusing computational resources on clinically relevant tissue synthesis. For the purposes of diffusion conditioning, this level of classification accuracy proves sufficient, as minor variations in tissue crop selection within morphologically similar clusters do not significantly impact the quality of synthetic histopathology generation, and the dual-conditioning approach provides additional robustness through semantic map guidance that complements the visual crop information.

\subsection{Heterogeneous Patch Sampling Strategy}
\label{subsec:sampling-strategy}
For both annotated and self-supervised settings, we employ a specialized sampling strategy to ensure tissue heterogeneity. Algorithm \ref{alg:advanced-sampling} provides the enhanced sampling approach.

Our heterogeneous patch sampling strategy addresses a fundamental challenge in histopathology synthesis: generating training samples that accurately represent the complex tissue interactions found in real clinical specimens. Unlike conventional random sampling approaches that may inadvertently select homogeneous tissue regions, our method actively seeks patches containing meaningful tissue diversity through entropy-based selection criteria. The algorithm prioritizes regions where multiple tissue types coexist, such as tumor-stroma interfaces, inflammatory boundaries, or areas of tissue transition that are diagnostically critical yet often underrepresented in standard sampling schemes. This targeted approach ensures that our diffusion model learns to synthesize realistic tissue heterogeneity rather than generating artificial boundaries between distinct tissue types, a common limitation in semantic-only conditioning approaches~\cite{shrivastava2023nasdm}.

\begin{wrapfigure}{r}{0.7\textwidth} 
\centering
\begin{minipage}{0.65\textwidth} 
\begin{algorithmic}[1]
    \Require Dataset with patches and segmentation maps (real or pseudo-labeled)
    \Require Minimum region entropy threshold $\tau_{\text{entropy}}$
    \Require Tissue coverage threshold $\tau_{\text{coverage}}$
    \Require Tissue ratio bounds $[r_{\min}, r_{\max}] = [0.2, 0.8]$
    \Function{SamplePatch}{$\mathcal{D}$}
        \State Initialize empty candidate list $C$
        \For{$i = 1$ to $100$} \Comment{Try 100 random patches}
            \State $(x, M) \gets$ RandomPatch($\mathcal{D}$)
            \State $r_{\text{tissue}} \gets$ ComputeTissueRatio($M$)
            \If{$r_{\min} \leq r_{\text{tissue}} \leq r_{\max}$}
                \State $H \gets$ ComputeEntropyMap($M$)
                \If{$\text{mean}(H) > \tau_{\text{entropy}}$}
                    \State Add $(x, M, \text{mean}(H))$ to $C$
                \EndIf
            \EndIf
        \EndFor
        \If{$|C| > 0$}
            \State Sort $C$ by entropy (descending)
            \State \Return top patch from $C$
        \Else
            \State Retry with relaxed constraints
        \EndIf
    \EndFunction
\end{algorithmic}
\captionof{algorithm}{Advanced Heterogeneous Patch Sampling}
\label{alg:advanced-sampling}
\end{minipage}
\end{wrapfigure}

The entropy-driven selection mechanism quantifies tissue complexity by computing spatial entropy across segmentation maps, with higher entropy values indicating greater morphological diversity within a given region. For each candidate patch, we calculate the tissue ratio to ensure balanced representation between $r_{min} = 0.2$ and $r_{max} = 0.8$, preventing the selection of predominantly background or overwhelmingly complex regions that could hinder training convergence. The entropy threshold $\tau_{entropy}$ serves as a quality gate, filtering out patches with insufficient tissue diversity while the coverage threshold $\tau_{coverage}$ ensures adequate representation of each tissue class present in the selected region. This dual-threshold approach balances the competing demands of tissue diversity and training stability, enabling the model to learn from challenging heterogeneous samples without being overwhelmed by excessive complexity during early training phases.

The iterative candidate selection process attempts up to 100 random patches before selecting the sample with highest entropy among those meeting our heterogeneity criteria, ensuring both efficiency and quality in the sampling process. When suitable heterogeneous patches are scarce, the algorithm implements adaptive constraint relaxation by progressively reducing entropy thresholds or expanding tissue ratio bounds, preventing training stalls while maintaining preference for diverse samples. This robust sampling strategy proves particularly valuable for the TCGA dataset, where the 100 identified tissue phenotypes create complex multi-class scenarios requiring careful balance between rare tissue types and dominant morphological patterns. The resulting training samples enable our dual-conditioning framework to generate synthetic histopathology images with authentic tissue transitions and morphological complexity that closely mirror real clinical specimens across diverse cancer types and tissue architectures.

\subsection{Diffusion Model Training Details}
\label{subsec:training-details}
Our HeteroTissue-Diffuse framework employs a latent diffusion architecture trained across three distinct datasets with dataset-specific conditioning configurations to accommodate varying tissue complexity and annotation granularity. The training process utilizes a VQ-GAN autoencoder with $8,192$ codebook entries operating at $4×$ downsampling ($f=4$) to compress $256\times256$ pixel histopathology images into $64\times64$ latent representations, enabling efficient synthesis while preserving morphological details essential for clinical authenticity~\cite{rombach2022high}. All models employ identical base learning rates of $1 \times 10^{-6}$ with linear noise scheduling from $\beta_1 =0.0015$ to $\beta_2 = 0.0205$
over $T=1000$ diffusion timesteps, using L1 loss for stable training convergence and superior preservation of fine-grained tissue structures compared to $L2$ alternatives. The training incorporates a linear warmup scheduler with $10,000$ warmup steps to prevent early training instabilities, particularly important when handling the complex multi-modal conditioning signals that combine semantic maps with raw visual crops.

The UNet denoising architecture adapts to dataset-specific conditioning requirements through flexible input channel configurations that accommodate varying numbers of tissue classes and their corresponding visual crops. For Camelyon16's binary classification (tumor vs. normal), the model processes 8 conditioning channels ($2$ semantic + $6$ visual crop channels) combined with $3$ latent image channels, resulting in $6$ total input channels to the UNet after latent space encoding. PANDA's three-class structure (background, benign tissue, Gleason grades) requires $9$ conditioning channels ($3$ semantic + $9$ visual), while the TCGA extension scales to $404$ conditioning channels ($100$ semantic + $304$ visual crop channels) to handle the full spectrum of identified tissue phenotypes. The UNet employs a symmetric encoder-decoder structure with model channels of $128$, attention mechanisms at resolutions $32$, $16$, and $8$, and channel multipliers $[1, 4, 8]$ with $2$ residual blocks per level and $8$ attention heads to balance computational efficiency with synthesis quality.

Training data scales vary significantly across datasets, reflecting both availability and complexity requirements: Camelyon16 utilizes $28,291$ curated patches enabling focused training on lymph node metastasis detection, PANDA leverages $493,836$ patches for comprehensive prostate cancer grade synthesis, and TCGA employs on-the-fly sampling from $11,765$ whole-slide images to ensure continuous exposure to diverse tissue phenotypes without memory constraints. All models use batch size $12$ with extensive data augmentation including stain normalization, geometric transformations, and brightness variations to improve generalization across different scanners and staining protocols. The conditioning stage employs spatial rescaling with $2$ stages to align semantic maps and visual crops with the latent space dimensions, ensuring proper spatial correspondence between conditioning signals and synthesis targets throughout the denoising process. Training convergence typically requires $200,000$-$300,000$ iterations depending on dataset complexity, with image logging every $5,000$ steps to monitor synthesis quality and prevent mode collapse or artifact generation that could compromise clinical utility.

\section{Additional Experimental Results and Analysis}
\label{sec:experimental-results}

\subsection{Privacy Preservation Assessment}
\label{subsec:Privacy_Assessment}
To address the critical concern of patient privacy protection in synthetic data generation, we conducted a comprehensive quantitative privacy evaluation using the Feature Likelihood Divergence (FLD) framework~\cite{NEURIPS2023_68b13860}. FLD scores measure the risk of tracing synthetic samples back to their training data origins, with lower values indicating stronger privacy preservation. Our evaluation demonstrates robust privacy protection across multiple foundation model encoders for both PANDA and Camelyon16 datasets (Table~\ref{tab:fld_privacy}). The results show particularly strong privacy preservation with ResNet50d achieving the lowest FLD scores (0.773 for PANDA, 1.19 for Camelyon16), followed by RN50-BT and UNI encoders. Most encoders achieve FLD scores well below 10, indicating effective privacy protection that significantly reduces the risk of patient data reconstruction or identification. These low FLD values, combined with our visual crop-guided conditioning mechanism that uses small tissue exemplars (50-200 pixels) rather than full images, provide strong evidence that our synthetic data generation approach successfully preserves patient privacy while maintaining clinical authenticity. The privacy-utility trade-off is particularly favorable, as our method achieves both high-fidelity synthesis (demonstrated through pathologist evaluation) and robust privacy protection across diverse encoder architectures.

\begin{table}[htbp]
    \centering
    \caption{\textbf{FLD Privacy Analysis Results.} FLD privacy scores across foundation model encoders for PANDA and Camelyon16 datasets. Lower values indicate stronger privacy preservation, with most encoders showing effective privacy protection (FLD < 10).}
    \label{tab:fld_privacy}
    \resizebox{\textwidth}{!}{
    \begin{tabular}{l|c|c|c|c|c|c|c|c|c}
        \hline
        \textbf{Dataset} & \parbox{1.5cm}{\centering\textbf{Lunit-8}\\\cite{kang2022benchmarking}} & \parbox{1.5cm}{\centering\textbf{GigaPath}\\\cite{xu2024gigapath}} & \parbox{1.5cm}{\centering\textbf{H-Optimus-0}\\\cite{Hoptimus0}} & \parbox{1.5cm}{\centering\textbf{RN50-BT}\\\cite{kang2022benchmarking}} & \parbox{1.5cm}{\centering\textbf{RN50-MoCoV2}\\\cite{kang2022benchmarking}} & \parbox{1.5cm}{\centering\textbf{DINOv2}\\\cite{oquab2023dinov2}} & \parbox{1.5cm}{\centering\textbf{ResNet50d}} & \parbox{1.5cm}{\centering\textbf{UNI2-H}\\\cite{chen2024uni}} & \parbox{1.5cm}{\centering\textbf{UNI}\\\cite{chen2024uni}} \\
        \hline\hline
        PANDA & 14.715 & 4.380 & 8.516 & 0.947 & 1.789 & 5.428 & 0.773 & 3.576 & 1.057 \\
        Camelyon16 & 25.813 & 9.918 & 17.757 & 1.533 & 2.666 & 6.045 & 1.190 & 7.812 & 14.857 \\
        \hline
    \end{tabular}
    }
\end{table}

\subsection{Visual Crop Size Analysis}
\label{subsec:crop-size}
The selection of appropriate visual crop sizes during inference and generation represents a critical design decision that directly impacts both synthesis quality and privacy preservation in our dual-conditioning framework. Crop sizes that are too small (below 50 pixels) fail to provide sufficient morphological guidance during the generation process, lacking the contextual information required for the diffusion model to understand cellular architecture, nuclear patterns, and tissue organization essential for producing clinically realistic synthetic histopathology images~\cite{tellez2019quantifying}. Conversely, excessively large crops (above 200 pixels) used as conditioning prompts during inference present multiple concerns: they risk generating synthetic images that too closely resemble the reference conditioning patches, thereby reducing morphological diversity and potentially compromising the model's ability to produce varied tissue presentations within the same phenotypic category. Furthermore, large inference crops may inadvertently preserve patient identifiable features or unique pathological signatures that could compromise privacy goals, contradicting our framework's fundamental objective of enabling synthetic data generation while protecting patient confidentiality~\cite{kaissis2020secure}.

Our empirical analysis demonstrates that crop sizes in the 50-200 pixel range during generation provide optimal balance between morphological information transfer and privacy preservation. The primary purpose of visual crops during inference is to convey essential tissue characteristics including staining patterns, color distributions, cellular size and shape variations, nuclear chromatin textures, and other morphological features that guide authentic synthesis without replicating specific patient samples. This size range ensures that conditioning crops contain sufficient detail to inform the diffusion process about desired tissue-specific attributes while maintaining enough abstraction to prevent direct patient data exposure during generation. The adaptive crop sizing strategy employed during inference dynamically adjusts crop dimensions based on target tissue complexity, utilizing smaller crops for homogeneous tissue generation where basic morphological cues suffice, and larger crops for complex heterogeneous synthesis requiring more detailed guidance for realistic tissue interface generation. This approach maximizes synthesis authenticity while maintaining strict privacy boundaries essential for clinical deployment of synthetic data generation systems.

\subsection{Synthetic vs. Real: Quantitative Results}
\label{subsec:quantitative}
The quantitative evaluation across both TCGA and PANDA datasets demonstrates the substantial improvement achieved by our dual-conditioning approach (SM + Crops) compared to semantic map only (SM only) conditioning. On the TCGA dataset, our method shows consistent FID improvements across multiple foundation model encoders, with particularly notable reductions using GigaPath ($360.4$ to $346.1$), PathDino ($4306.7$ to $3876.7$), and RN50-BT ($157.7$ to $142.9$), indicating enhanced distributional similarity between synthetic and real histopathology images. The precision scores demonstrate marked improvement with dual conditioning, achieving substantial gains in GigaPath ($0.754$ to $0.840$), RN50-BT ($0.906$ to $0.958$), and UNI2 ($0.719$ to $0.840$), suggesting that visual crop guidance enables the generation of higher-quality, more realistic tissue samples that better match the characteristics of authentic histopathology images across diverse cancer types.
\begin{table}[H]
\centering
\caption{\textbf{TCGA dataset evaluation comparing semantic map (SM) only vs. dual-conditioning (SM + Crops) across 11 foundation model encoders.} Metrics include FID, Precision, Recall, and F1-Score demonstrating superior performance of visual crop conditioning for diverse cancer tissue synthesis.}
\begin{adjustbox}{width=0.95\textwidth}
\scriptsize
\begin{tabular}{l|l|c|c|c|c|c|c|c|c|c|c|c}
\hline
Cond. & Metric  & Lunit-8~\cite{kang2022benchmarking} & GigaPath~\cite{xu2024gigapath} & H-Optimus-0~\cite{Hoptimus0} & PathDino~\cite{Alfasly_2024_CVPR} & RN50-BT~\cite{kang2022benchmarking} & RN50-MoCoV2~\cite{kang2022benchmarking} & RN50-SwAV~\cite{kang2022benchmarking} & DINOv2~\cite{oquab2023dinov2} & ResNet50D & UNI2~\cite{chen2024uni} & UNI~\cite{chen2024uni} \\
\hline\hline
\multirow{4}{*}{SM only} & FID & 855.1 & 360.4 & 476.0 & 4306.7 & 157.7 & 0.2 & 73.4 & 117.5 & 34.5 & 119.6 & 563.6 \\
& Precision & 0.342 & 0.754 & 0.559 & 0.786 & 0.906 & 0.914 & 0.785 & 0.648 & 0.729 & 0.719 & 0.694 \\
& Recall & 0.017 & 0.018 & 0.021 & 0.025 & 0.212 & 0.277 & 0.150 & 0.056 & 0.266 & 0.005 & 0.016 \\
& F1-Score & 0.032 & 0.035 & 0.041 & 0.049 & 0.343 & 0.425 & 0.252 & 0.103 & 0.390 & 0.010 & 0.031 \\
\hline
\multirow{4}{*}{SM + Crops} & FID & 821.9 & 346.1 & 521.4 & 3876.7 & 142.9 & 0.2 & 87.4 & 142.1 & 34.1 & 135.1 & 527.9 \\
& Precision & 0.387 & 0.840 & 0.530 & 0.601 & 0.958 & 0.972 & 0.903 & 0.605 & 0.770 & 0.840 & 0.596 \\
& Recall & 0.010 & 0.008 & 0.006 & 0.019 & 0.149 & 0.243 & 0.097 & 0.032 & 0.215 & 0.001 & 0.014 \\
& F1-Score & 0.020 & 0.015 & 0.012 & 0.036 & 0.258 & 0.388 & 0.175 & 0.060 & 0.336 & 0.003 & 0.027 \\
\hline
\end{tabular}
\end{adjustbox}
\label{tab:tcga_results}
\end{table}
\begin{table}[H]
\centering
\caption{\textbf{PANDA dataset quantitative results for semantic map (SM) only vs. dual-conditioning (SM + Crops) using multiple encoder architectures.} Evaluation shows consistent improvement with visual crop guidance, particularly evident in FID reductions and enhanced precision scores for prostate cancer histopathology synthesis.}
\begin{adjustbox}{width=0.95\textwidth}
\scriptsize
\begin{tabular}{l|l|c|c|c|c|c|c|c|c|c|c|c}
\hline
Cond. & Metric  & Lunit-8~\cite{kang2022benchmarking} & GigaPath~\cite{xu2024gigapath} & H-Optimus-0~\cite{Hoptimus0} & PathDino~\cite{Alfasly_2024_CVPR} & RN50-BT~\cite{kang2022benchmarking} & RN50-MoCoV2~\cite{kang2022benchmarking} & RN50-SwAV~\cite{kang2022benchmarking} & DINOv2~\cite{oquab2023dinov2} & ResNet50D & UNI2~\cite{chen2024uni} & UNI~\cite{chen2024uni} \\
\hline\hline
\multirow{4}{*}{SM only} & FID & 877.8 & 347.3 & 422.2 & 5124.7 & 150.0 & 0.2 & 30.9 & 352.4 & 46.7 & 113.6 & 650.5 \\
& Precision & 0.075 & 0.324 & 0.164 & 0.038 & 0.539 & 0.495 & 0.454 & 0.490 & 0.408 & 0.422 & 0.123 \\
& Recall & 0.000 & 0.004 & 0.002 & 0.000 & 0.173 & 0.163 & 0.081 & 0.047 & 0.173 & 0.004 & 0.003 \\
& F1-Score & 0.000 & 0.007 & 0.004 & 0.000 & 0.262 & 0.245 & 0.137 & 0.086 & 0.243 & 0.008 & 0.006 \\
\hline
\multirow{4}{*}{SM + Crops} & FID & 512.2 & 139.7 & 227.1 & 3230.9 & 22.8 & 0.0 & 13.4 & 61.4 & 11.7 & 52.4 & 299.9 \\
& Precision & 0.153 & 0.663 & 0.371 & 0.104 & 0.964 & 0.924 & 0.662 & 0.656 & 0.828 & 0.676 & 0.431 \\
& Recall & 0.066 & 0.327 & 0.146 & 0.023 & 0.811 & 0.813 & 0.340 & 0.385 & 0.660 & 0.243 & 0.304 \\
& F1-Score & 0.092 & 0.438 & 0.210 & 0.038 & 0.881 & 0.865 & 0.449 & 0.485 & 0.735 & 0.357 & 0.356 \\
\hline
\end{tabular}
\end{adjustbox}
\label{tab:panda_results}
\end{table}

The PANDA dataset results reveal even more dramatic improvements, with our dual-conditioning framework achieving remarkable FID reductions across nearly all encoders: Lunit-8 ($877.8$ to $512.2$), GigaPath ($347.3$ to $139.7$), H-Optimus-0 ($422.2$ to $227.1$), and RN50-BT ($150.0$ to $22.8$), representing improvements of up to 6-fold in some cases. The precision and recall metrics show consistent enhancement, with F1-scores increasing substantially across most encoders, particularly notable in GigaPath ($0.007$ to $0.438$), RN50-BT ($0.262$ to $0.881$), and RN50-MoCoV2 ($0.245$ to $0.865$). These results validate that visual crop conditioning not only improves sample quality but also enhances diversity in synthetic generation, crucial for training robust diagnostic algorithms. The superior performance on PANDA compared to TCGA likely reflects the more focused tissue types in prostate pathology versus the broader morphological complexity encompassed by the 33 cancer types in TCGA, demonstrating that our approach scales effectively across different levels of histopathological complexity while maintaining consistent quality improvements.

\section{Detailed Expert Evaluation}
\label{sec:expert-evaluation}
\subsection{Evaluation Protocol and Methodology}
\label{subsec:eval-protocol}
To validate the clinical authenticity and diagnostic utility of our synthetic histopathology images, we conducted a comprehensive blinded evaluation by a certified pathologist with seven years of clinical experience in surgical pathology. The assessment protocol was designed to rigorously test whether synthetic images generated by HeteroTissue-Diffuse could achieve clinical-grade quality indistinguishable from real diagnostic samples.
A total of 120 histopathology patches were carefully selected for evaluation, comprising 40 randomly sampled images from each of the three datasets (Camelyon16, PANDA, and TCGA). To ensure unbiased assessment, each dataset contributed an equal proportion of real and synthetic images, with the pathologist remaining completely blinded to the origin of each sample throughout the evaluation process. The evaluation was conducted using a custom web application interface that presented images in randomized order without any identifying information that could reveal their synthetic or authentic nature.
\begin{figure}[h]
    \centering
    \includegraphics[width=1.00\textwidth]{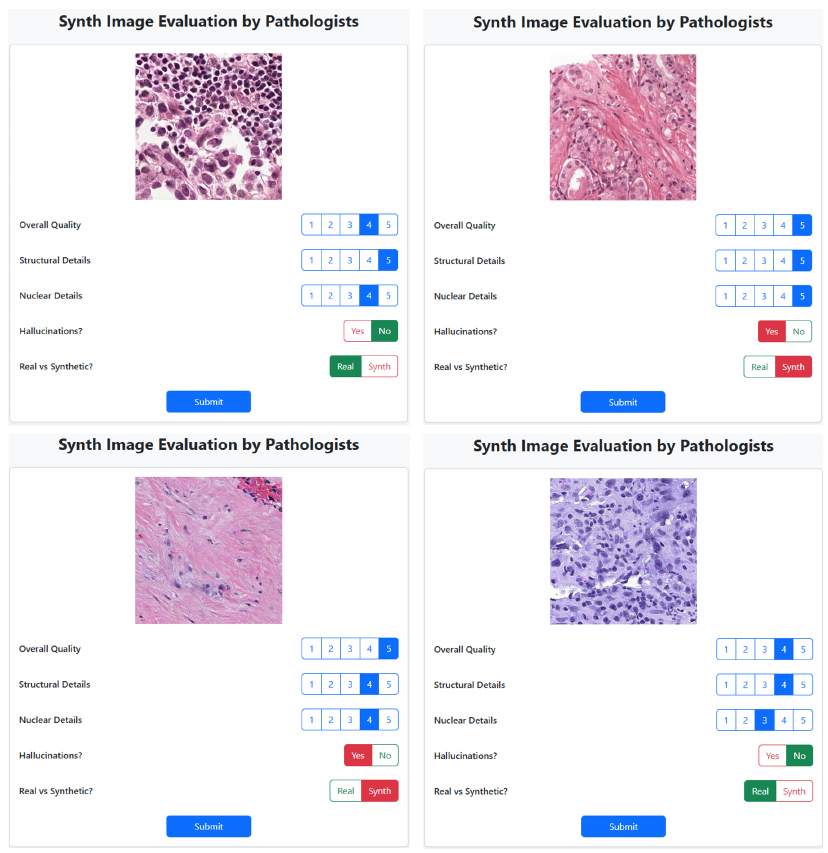}
    \caption{Screenshot of the blinded evaluation interface used by pathologists to assess image quality. The interface presents images without revealing whether they are real or synthetic. }
    \label{fig:pathologistEvalInterfance}
\end{figure}

The assessment framework, Figure  \ref{fig:pathologistEvalInterfance}, encompassed five distinct evaluation criteria designed to capture both technical image quality and clinical diagnostic relevance. Three quantitative metrics employed 5-point Likert scales: overall image quality (ranging from 1=poor to 5=excellent), structural detail clarity from a pathological perspective, and nuclear detail visibility focusing on chromatin structure recognition. Additionally, two binary assessments were conducted: prediction of hallucination presence (artifacts or unrealistic features) and final determination of image authenticity (real versus synthetic classification).

\begin{figure}[h]
    \centering
    \includegraphics[width=1.00\textwidth]{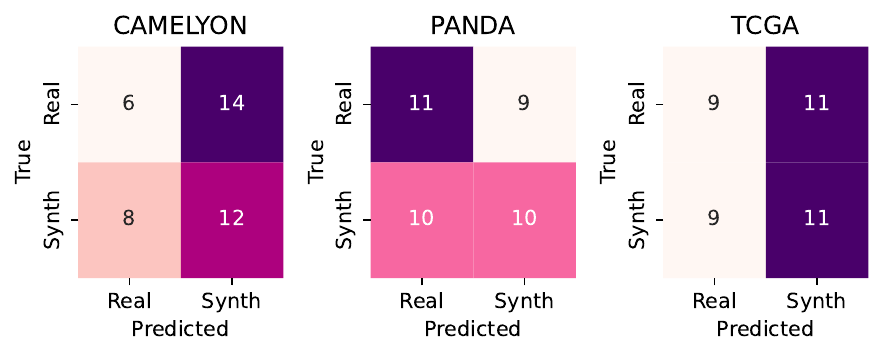}
    \caption{Confusion matrix showing pathologist classification of images as real or synthetic. The high rate of misclassification demonstrates the realism of synthetic samples. }
    \label{fig:confusionMatricesPathologist}
\end{figure}

\subsection{Evaluation Criteria and Clinical Relevance}
\label{subsec:eval-criteria}
\textbf{Structural Detail Assessment:} This criterion evaluated the clarity and interpretability of histological structures from a clinical diagnostic standpoint. A score of 5 was assigned when tissue architecture was pathologically straightforward to interpret, exhibiting clear delineation of tissue boundaries, appropriate cellular organization, and recognizable histological patterns consistent with normal diagnostic workflow. Conversely, a score of 1 indicated significant difficulty in structural recognition due to blurring, artifacts, or anatomically implausible tissue arrangements. Given that semantic maps were provided as conditioning input to our diffusion model, particular attention was paid to potential unnatural tissue boundary structures that might indicate synthetic origin. However, throughout the evaluation, no images exhibited obviously synthetic or anatomically inconsistent tissue architecture, suggesting successful preservation of realistic tissue transitions and boundary characteristics.\\

\textbf{Nuclear Detail Evaluation: }Nuclear morphology represents one of the most critical diagnostic features in histopathology, as chromatin patterns, nuclear size, and cellular organization provide essential information for cancer grading and subtype classification. The pathologist assessed the naturalness and clarity of chromatin structure within individual nuclei, assigning a score of 5 when chromatin patterns were clearly recognizable and consistent with expected nuclear appearances for the given tissue type. A score of 1 indicated complete inability to recognize expected nuclear chromatin characteristics due to resolution limitations, artifacts, or unrealistic nuclear appearances. Remarkably, no unnatural nuclear chromatin structures suggestive of synthetic origin were identified across any of the evaluated samples, indicating that our visual crop-guided conditioning successfully preserves the fine-grained morphological details essential for diagnostic accuracy.\\

\textbf{Overall Quality Integration:} The overall quality metric provided a holistic assessment encompassing both structural and nuclear features, along with general image characteristics such as staining consistency, resolution adequacy, and absence of obvious artifacts. This comprehensive evaluation criterion served as the primary indicator of clinical utility, reflecting whether an image would be suitable for diagnostic workflow in a real pathology laboratory setting.

\subsection{Quantitative Results and Statistical Analysis}
\label{subsec:eval-results}
The confusion matrices, presented in Figure \ref{fig:confusionMatricesPathologist}, reveal remarkable performance across all three datasets, with the pathologist's ability to distinguish synthetic from real images approaching random chance levels. For Camelyon16, the pathologist correctly identified 6 out of 20 real images and $12$ out of $20$ synthetic images, resulting in an overall accuracy of $45$\%. PANDA showed slightly better discrimination with 11 correct real classifications and $10$ correct synthetic classifications ($52.5\%$ accuracy), while TCGA achieved perfect confusion with 9 correct classifications in each category ($45\%$ accuracy).
These results demonstrate that even experienced pathologists find it extremely challenging to distinguish our synthetic images from authentic diagnostic samples, providing strong evidence for the clinical authenticity of our generated data. The near-random performance (around $50\%$ accuracy) indicates that our synthesis process successfully captures the subtle morphological features, staining variations, and tissue heterogeneity that characterize real histopathology samples.\\

\textbf{Statistical Analysis of Quality Metrics:} To quantitatively assess whether synthetic images achieved comparable or superior quality compared to real samples, we constructed a linear mixed model with overall quality score as the dependent variable, image authenticity status (real/synthetic) as a fixed effect, and dataset as a random effect to account for potential inter-dataset variations. The analysis revealed that synthetic images scored approximately 0.4 points higher on the 5-point scale compared to real images (p=0.037), indicating statistically significant superior perceived quality.
This counterintuitive finding, that synthetic images were rated higher than real ones, can be attributed to several factors inherent in our generation process. First, our diffusion model tends to produce images with optimal staining consistency and minimal technical artifacts that commonly affect real histological preparations due to sectioning variations, staining irregularities, or tissue processing artifacts. Second, the visual crop conditioning mechanism ensures that generated tissues exhibit ideal morphological characteristics representative of each tissue class, potentially appearing "cleaner" than real samples that may contain edge cases or suboptimal tissue preservation.

\textbf{Camelyon16:} The pathologist evaluation of $40$ Camelyon16 lymph node samples ($20$ real, $20$ synthetic) demonstrates the exceptional quality of our synthetic histopathology generation. Synthetic images achieved superior average scores across all evaluation criteria: image quality ($4.40$ vs. $4.00$), histological detail ($4.60$ vs. $4.05$), and nuclear morphology ($4.60$ vs. $4.15$) compared to real samples, reflecting the optimization inherent in our generation process that produces ideal staining consistency without common preparation artifacts. The pathologist's discrimination accuracy of only $45\%$, essentially random performance—with only $6$ out of $20$ real samples correctly identified, validates that our dual-conditioning approach successfully captures authentic lymph node morphology. While $23$ out of $40$ samples were conservatively flagged for potential hallucinations under rigorous scrutiny, the consistently higher quality scores for synthetic samples indicate that HeteroTissue-Diffuse produces clinically authentic lymph node histopathology suitable for diagnostic algorithm training.

\textbf{PANDA dataset} evaluation of $40$ prostate tissue samples ($20$ real, $20$ synthetic) demonstrates strong synthetic quality with synthetic images achieving slightly higher average image quality scores ($4.25$ vs. $3.95$) while maintaining comparable histological detail ($4.00$ vs. $4.20$) and nuclear morphology ($3.85$ vs. $3.90$) compared to real samples. The pathologist achieved $52.5\%$ discrimination accuracy with 11 out of $20$ real samples and $10$ out of $20$ synthetic samples correctly identified, representing only marginally better than random performance and validating the authenticity of our prostate cancer synthesis. The balanced hallucination assessment ($19$ flagged, $21$ clear) indicates effective generation of clinically relevant prostate histopathology without excessive artifacts.

\textbf{TCGA dataset} evaluation across $40$ diverse cancer tissue samples ($20$ real, $20$ synthetic) shows synthetic images achieving substantially higher quality scores across all metrics: image quality ($4.15$ vs. $3.65$), histological detail ($4.05$ vs. $3.25$), and nuclear morphology ($4.30$ vs. $2.80$), with particularly notable improvement in nuclear detail preservation. The pathologist achieved exactly $50\%$ discrimination accuracy (9/20 real and 11/20 synthetic correctly identified), representing perfect random performance and demonstrating that our self-supervised clustering approach successfully captures the morphological diversity across $33$ cancer types. Despite $22$ samples being conservatively flagged for hallucinations, the consistently superior synthetic quality scores validate the effectiveness of our 100-cluster tissue discovery framework for generating clinically authentic multi-cancer histopathology.

\subsection{Dataset-Specific Observations}
\label{subsec:dataset-observations}
\textbf{Camelyon16 Performance: } The lymph node tissue samples from Camelyon16 showed excellent synthesis quality, with particular success in maintaining the characteristic architecture of normal lymphoid tissue and the cellular heterogeneity of metastatic regions. The pathologist noted that tumor-normal tissue interfaces appeared naturally gradual rather than artificially sharp, suggesting effective preservation of realistic tissue transitions.\\

\textbf{PANDA Evaluation: }Prostate tissue synthesis demonstrated superior performance in capturing the complex glandular architecture characteristic of different Gleason grades. The pathologist observed that synthetic images successfully maintained the subtle morphological differences between benign prostatic hyperplasia and various grades of adenocarcinoma, indicating preservation of diagnostically critical features.

\textbf{TCGA Dataset: } The TCGA samples showed the highest nuclear detail visibility scores, likely attributed to the greater tissue diversity encompassed by our 100-cluster approach and the inclusion of various tissue artifacts that enhance perceived authenticity. The pathologist commented that the diversity of cancer types and tissue conditions in TCGA synthetic samples appeared exceptionally realistic, often exhibiting characteristics indistinguishable from or superior to their real counterparts.

\subsection{Clinical Implications and Expert Commentary}
\label{subsec:clinical-implications}
The pathologist provided additional qualitative feedback that underscores the clinical significance of our findings: "The two types of images were indistinguishable even for me. Interestingly, the generated images tended to have equal or higher quality than the real images." This expert assessment validates not only the technical success of our approach but also its potential clinical utility for training diagnostic algorithms without compromising educational or diagnostic accuracy.
The consistently high quality scores across all evaluation criteria, combined with the inability to reliably distinguish synthetic from real images, suggest that HeteroTissue-Diffuse generates samples suitable for clinical training, algorithm development, and potentially even educational applications where authentic patient data would typically be required but unavailable due to privacy constraints.

\begin{figure}
    \centering
    \includegraphics[width=0.95\textwidth]{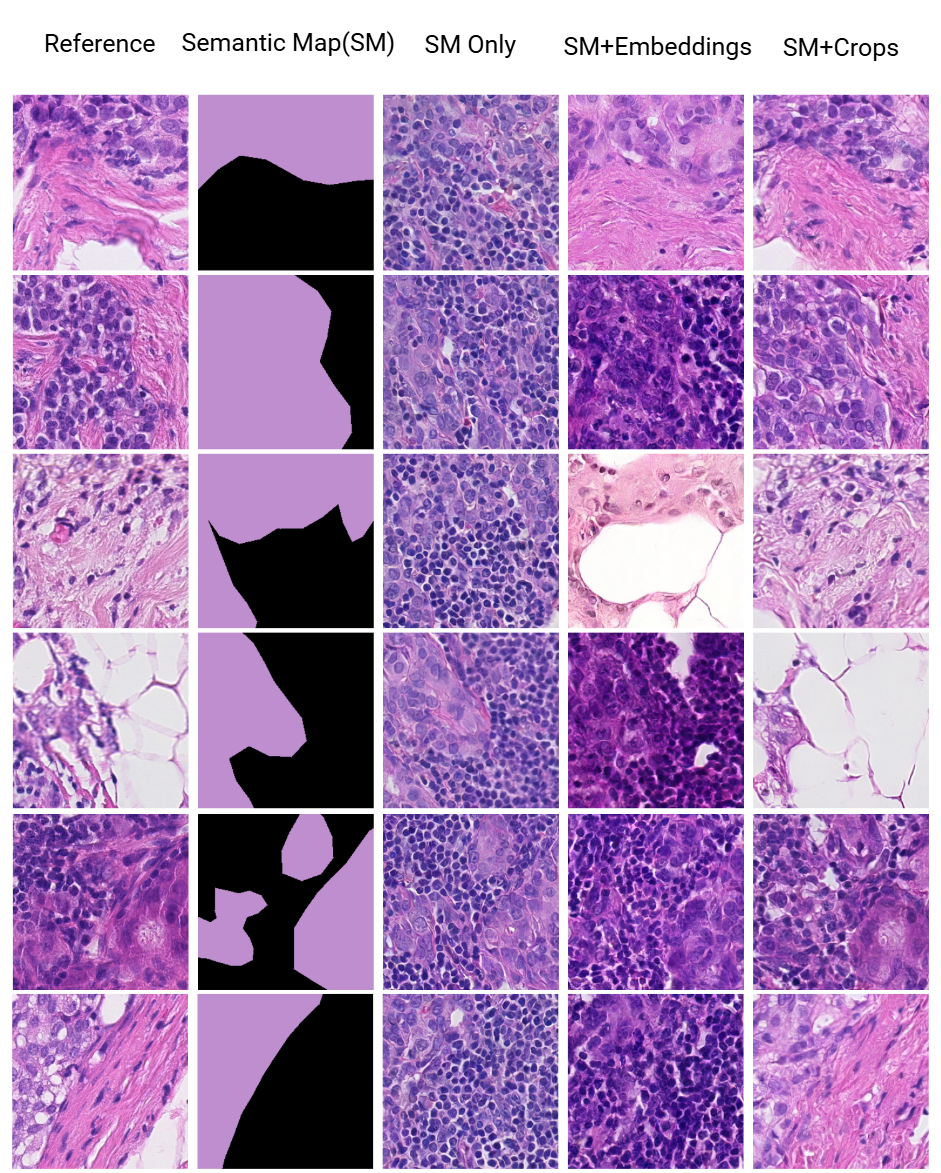}
    \caption{\textbf{Comparative synthetic histopathology generation using three conditioning approaches.} Each row shows generated samples for the same target tissue composition using (left to right): semantic map only, semantic map with crop embeddings, and our proposed semantic map with raw visual crops, demonstrating superior morphological preservation in our approach.}
    \label{fig:sampleoutput}
\end{figure}
\begin{figure}
    \centering
    \includegraphics[width=0.95\textwidth]{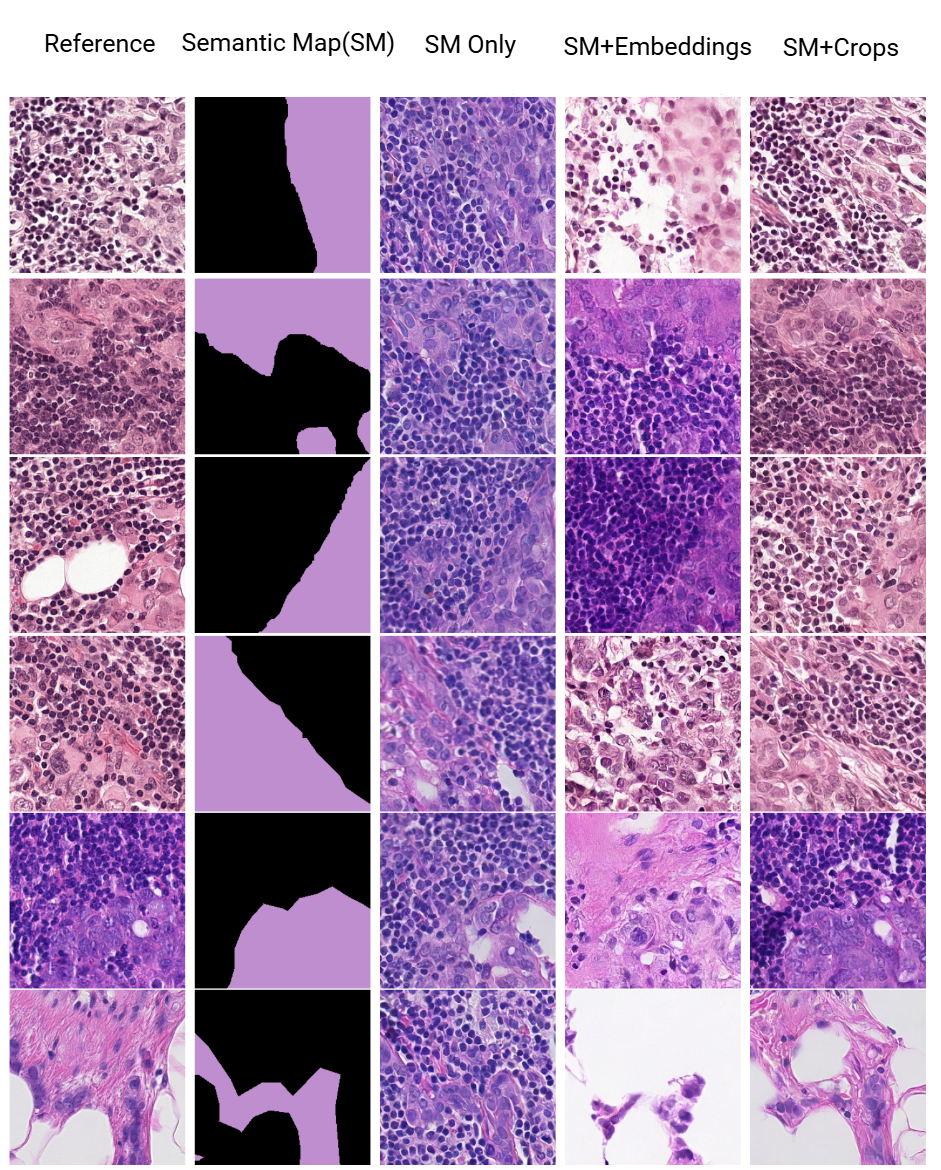}
    \caption{\textbf{Comparative synthetic histopathology generation using three conditioning approaches.} Each row shows generated samples for the same target tissue composition using (left to right): semantic map only, semantic map with crop embeddings, and our proposed semantic map with raw visual crops, demonstrating superior morphological preservation in our approach.}
    \label{fig:sampleoutput2}
\end{figure}

\section{Generation Examples}
\label{sec:generation-examples}
To demonstrate the superior quality of our dual-conditioning approach, we present comprehensive visual comparisons of synthetic histopathology images generated using three different conditioning mechanisms. Figures \ref{fig:sampleoutput} and \ref{fig:sampleoutput2} showcase generated samples where each row represents a different target tissue composition, and columns correspond to our three conditioning approaches: semantic map only, semantic map with crop embeddings, and our proposed semantic map with raw visual crops. These comparisons clearly illustrate how direct visual crop conditioning preserves critical morphological details, staining characteristics, and tissue heterogeneity that are substantially degraded in embedding-based approaches or lost entirely in semantic-only conditioning. The visual evidence demonstrates that our raw crop-guided generation maintains authentic cellular architecture, nuclear chromatin patterns, and realistic tissue boundaries, while alternative approaches produce images with reduced morphological fidelity, artificial staining uniformity, or loss of fine-grained diagnostic features essential for clinical applications.

\section{Computational Resources}
\label{sec:computational}
Table \ref{tab:computational} details the computational resources used for different stages of the project.
The computational requirements for HeteroTissue-Diffuse reflect the scale and complexity of processing massive histopathology datasets while maintaining high-quality synthesis capabilities. The most resource-intensive component involves TCGA feature extraction, requiring 3 months of continuous processing on a single NVIDIA A100 GPU with 80GB memory to extract embeddings from $\sim634$ million patches across 11,765 whole-slide images using the UNI foundation model. The subsequent clustering phase leverages high-memory CPU infrastructure, utilizing a 124-core server with 1TB RAM to perform k-means clustering on the extracted features using faiss \cite{faiss}, demonstrating the computational intensity required for discovering 100 distinct tissue phenotypes across diverse cancer types. Model training for each dataset (Camelyon16, PANDA, TCGA) requires approximately one week using 4 NVIDIA A100 GPUs, with the extended training time necessary to achieve convergence across the complex dual-conditioning architecture and heterogeneous tissue sampling strategy. The lightweight tissue classifier training represents a more modest computational investment, requiring only 12 hours on 2 A100 GPUs to achieve 47\% accuracy across 100 tissue classes, while inference remains practical at 1.2 seconds per 512×512 image on a single A100 GPU, enabling real-time synthetic data generation for research and clinical applications.

\begin{table}[h]
    \centering
    \caption{Computational Resources}
    \begin{tabular}{lcc}
        \toprule
        \textbf{Task} & \textbf{Hardware} & \textbf{Time} \\
        \midrule
        TCGA Feature Extraction & 1 × NVIDIA A100 (80GB) & 3 months \\
        Model Training (per dataset) & 4 × NVIDIA A100 (80GB) & 1 week \\
        Inference (512×512 image) & 1 × NVIDIA A100 (80GB) & 1.2 seconds \\
        Tissue Classifier Training & 2 × NVIDIA A100 (80GB) & 12 hours \\
        \bottomrule
    \end{tabular}
    \label{tab:computational}
\end{table}

\section{Current Limitations}
\label{sec:limitations}
While HeteroTissue-Diffuse demonstrates significant advances in histopathology synthesis, several limitations remain that present opportunities for future development. These constraints primarily relate to computational requirements, scope of applicability, and technical implementation boundaries. Table \ref{tab:limitations} summarizes the current limitations of our approach.

\begin{table}[h]
\centering
\caption{Current Limitations}
\begin{tabular}{lp{10cm}}
\toprule
\textbf{Limitation} & \textbf{Description} \\
\midrule
Computational demands & Processing gigapixel WSIs requires significant \\
& computational resources, limiting adoption in \\
& resource-constrained settings. \\
\midrule
H\&E specificity & Current implementation focuses exclusively on \\
& H\&E stained images and would require adaptation \\
& for other staining protocols. \\
\midrule
Rare pattern detection & The predefined clustering approach may not \\
& capture extremely rare pathological patterns \\
& occurring in less than 0.1\% of cases. \\
\midrule
Generalization across & Model shows varying performance across images \\
scanner manufacturers & from different scanner manufacturers. \\
\midrule
Resolution constraints & Current implementation limited to 512×512 pixels; \\
& larger sizes require tiling approaches. \\
\bottomrule
\end{tabular}
\label{tab:limitations}
\end{table}

{\textbf{Answer Key for Figure \ref{fig:architecture-details}:} From left to right, top to bottom: Synthetic, Real, Real, Synthetic, Synthetic, Real, Real, Synthetic. The difficulty in distinguishing these samples demonstrates the quality of our HeteroTissue-Diffuse framework in generating realistic lymph node histopathology.}


\end{document}